\documentclass{article}

\usepackage{microtype}
\usepackage{graphicx}
\usepackage{subcaption}
\usepackage{booktabs} 
\usepackage{calc} 
\usepackage{xcolor}
\usepackage[hidelinks]{hyperref}
\usepackage{multirow}
\usepackage{algorithm}
\usepackage{algorithmic}

\usepackage{paracol}
\usepackage{caption}
\usepackage{subcaption}

\usepackage[main, final]{neurips_2026}

\usepackage{amsmath}
\usepackage{amssymb}
\usepackage{mathtools}
\usepackage{amsthm}
\usepackage{physics}
\usepackage{tikz}
\usetikzlibrary{positioning,arrows.meta,calc,backgrounds,decorations.markings,decorations.text}
\usepackage{pgfplots}
\pgfplotsset{compat=1.18}
\usepgfplotslibrary{colormaps}
\usepackage[capitalize,noabbrev]{cleveref}

\theoremstyle{plain}

\theoremstyle{definition}

\theoremstyle{remark}

\usepackage{mdframed}
\usepackage{xcolor}
\usepackage{pifont}
\usepackage{makecell}
\usepackage{wrapfig}

\newcommand{\cmark}{\textcolor{green!60!black}{\ding{51}}}
\newcommand{\xmark}{\textcolor{red}{\ding{55}}}
\DeclareRobustCommand{\uncertain}{%
  \begin{tikzpicture}[baseline=-0.6ex]
    \draw[line width=0.9pt, draw=yellow!70!black] (0,0) circle (0.65ex);
  \end{tikzpicture}%
}
\newmdenv[
  backgroundcolor=gray!10,
  linecolor=gray!70,
  linewidth=1pt,
  roundcorner=5pt,
  skipabove=10pt,
  skipbelow=10pt
]{highlightbox}

\title{Show Me What You Don’t Know: Efficient Sampling from Invariant Sets for Model Validation}

\author{%
  Armand Rousselot \\
  Computer Vision and Learning Lab \\ Heidelberg University \\
  \texttt{armand.rousselot@iwr.uni-heidelberg.de}
  \And
  Joran Wendebourg \\
  Computer Vision and Learning Lab \\ Heidelberg University \\
  \And
  Ullrich Köthe \\
  Computer Vision and Learning Lab \\ Heidelberg University \\
}

\begin{document}

\maketitle

\begin{abstract}

The performance of machine learning models depends on discovering features in the data that are relevant to solving the given task while ignoring irrelevant variation. To check and visualize whether this is the case, we propose ‘invariance auditing’, a diagnostic tool to analyze feature extractors. Invariance auditing generates diverse inputs which all yield the same features as a given query input, and are therefore treated as identical by the model. Using external knowledge about the task at hand, this enables assessing if the learned invariances do or do not conform to the task's required semantics. Unlike existing work where a dedicated generative model is trained for each feature extractor, our algorithm is training-free and exploits a pretrained diffusion or flow-matching model to sample invariant inputs. Our fiber loss -- which penalizes feature mismatch -- guides the denoising process so the output shares the same representation as the query input. This replaces days of training with a single guided generation procedure at comparable quality. Experiments on popular datasets 
and model types 
demonstrate that our auditing method reveals invariances spanning very desirable and concerning behavior. For instance, it detects cases where Qwen-2B places patients with situs inversus (heart on the right side) in the same fiber as typical anatomy. Our code is available at \url{https://github.com/vislearn/InvarianceAuditing}.

\end{abstract}

\section{Introduction}
\label{sec:introduction}

Deep representation learning has seen massive success across applications \citep{bengio2013representation}, yet understanding and explaining learned representations remains an open problem \citep{doshi2017towards, samek2017explainable}.
The most prominent avenues strive to assign semantic meaning to the learned features \citep{ribeiro2016should, yang2022explaining, cunningham2023sparse} or invert feature representations of given models, in order to inspect which information they encode \citep{har2025visual, mahendran2015understanding}. 
However, to fully understand a learned representation, it is equally important to analyze what it does {\em not} encode. We refer to diagnostic tools for this purpose as {\em invariance auditing}, which reveals which features of the data a model ignores, surfacing systematic biases, shortcuts, and failure modes in its representations \citep{sammani2023visualizing, bordes2021high, rombach2020making}. Given a fixed model, the aim is to generate diverse natural inputs that are all assigned the same representation. Invariance auditing complements the analysis of adversarial robustness \citep{goodfellow2015explaining}: while the latter asks if very similar inputs can have vastly different representations, we search for very different inputs that have the same representations. 
\begin{figure}[h]
    \centering
    \begin{subfigure}[b]{0.45\linewidth}
    \includegraphics[width=\linewidth]{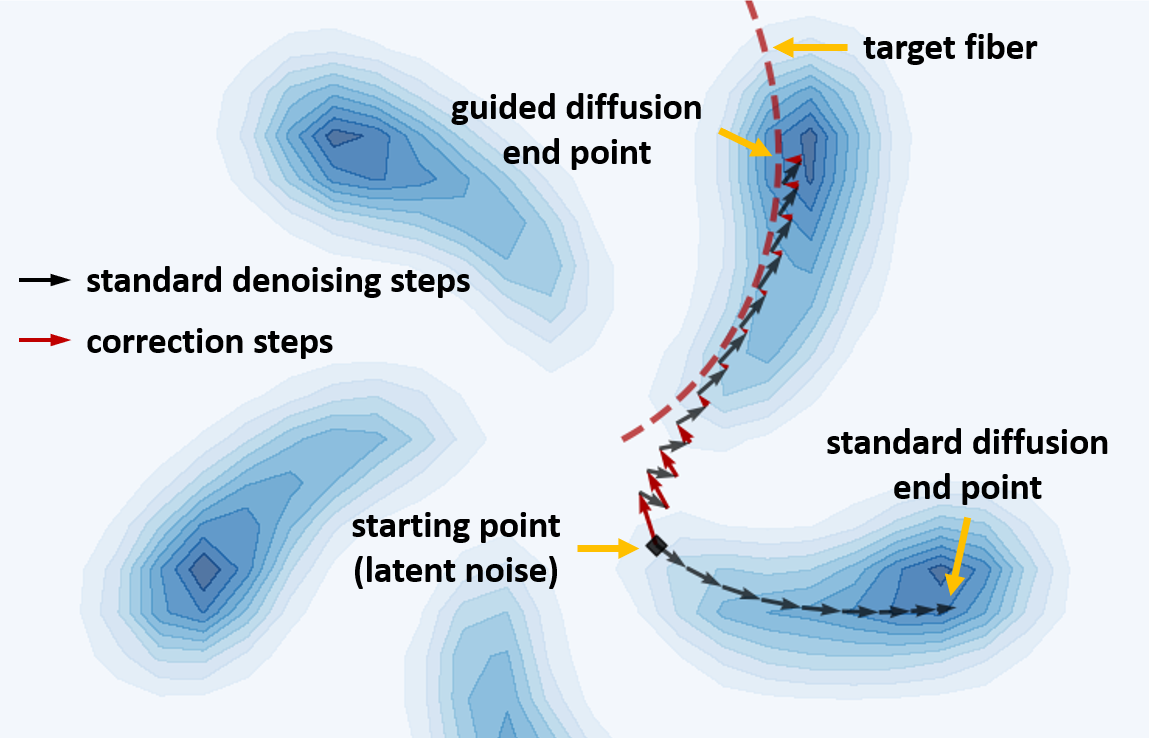}
    \caption{Black arrows represent the ordinary denoising steps in a flow-matching model. Alternating black (FM) and red (correction) arrows indicate the guided trajectory to the dashed target fiber.}
    \end{subfigure}
    \hspace{0.03\linewidth}
    \begin{subfigure}[b]{0.5\linewidth}
    \includegraphics[width=\linewidth]{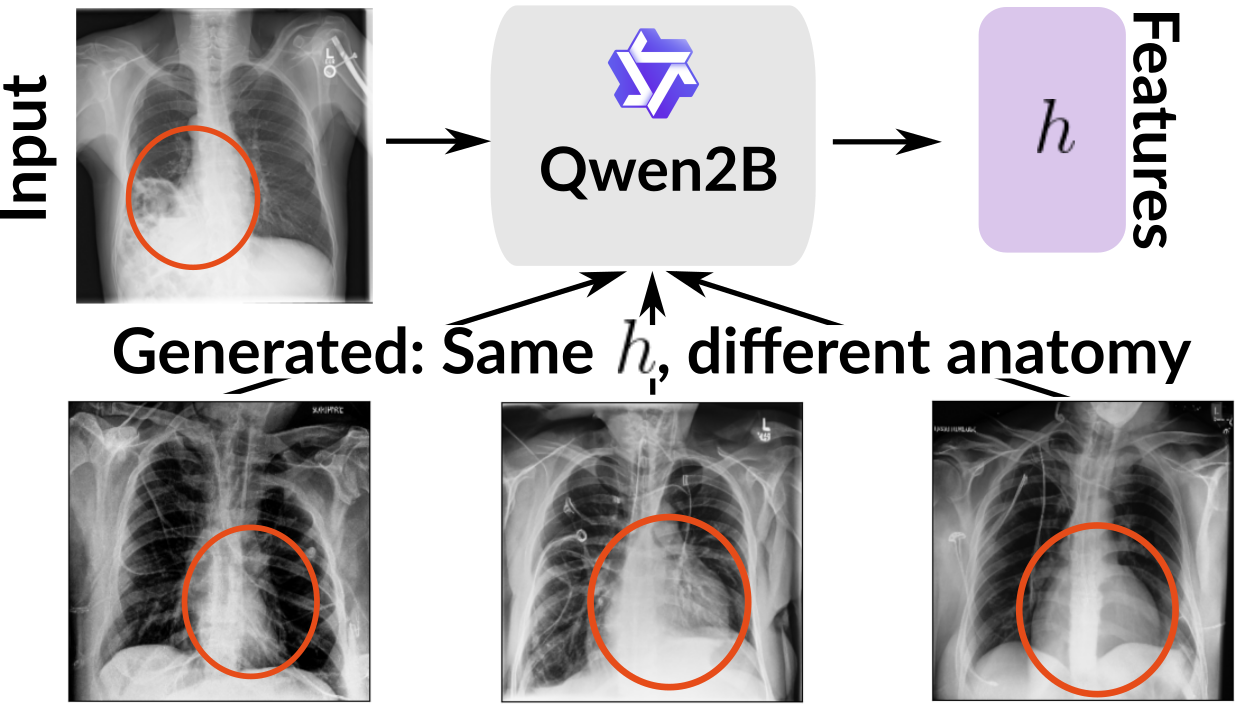}
    \caption{Given a case of "situs inversus", where the heart is located on the right side of the body, our invariance sampling -- guided by the Qwen-2B vision model -- finds several images that yield the same features but have the heart on the left side.}
    \end{subfigure}
    \caption{Illustration of invariance auditing using guided pretrained generative models.}
    \label{fig:Figure1}
    \vspace{-0.4cm}
\end{figure}

The \textit{fiber} (or invariant set) of the model $\phi$ to be analyzed (called the {\em subject model}) at representation value $h$ is the equivalence class defined as the set of inputs $x$, that share the representation $\phi(x) = h$. The pioneering work of \citet{rombach2020making} formulated invariance auditing as training a conditional model that samples from the fiber distribution $p(x|h)$ (called the {\em fiber model}). While several works have since built on this approach \citep{bordes2021high, eulig2024reconstructing} it comes with two major drawbacks: For every subject model a new fiber model has to be trained entirely from scratch, and there is no control over how closely samples lie to the target fiber.

We address these challenges in two complementary ways.
First we propose to leverage the difference between sample and target representation, which we call \textit{fiber loss}, as a regularizer for fiber model training. It was first introduced as a standalone optimization objective in \citet{mahendran2015understanding} and we show that fiber models trained with fiber loss are more reliable in terms of their achieved level of invariance. To this end we construct a benchmark based on color MNIST from \citet{rombach2020making} where the invariant sets and the distributions on them are known exactly. Second, we propose a training-free method for invariance auditing. Rather than retraining fiber models, we steer the denoising path of a pretrained diffusion/flow-matching model using the fiber loss to penalize feature mismatch (\cref{fig:Figure1}).
\begin{wraptable}{l}{0.6\textwidth}
    \centering
    \caption{Comparison of different methods for invariance auditing. We give a ``\cmark'' if a method fulfills a property and a ``\uncertain'' if a property is approximately fulfilled.}
    \resizebox{\linewidth}{!}{
        
\begin{tabular}{lcccc}
    \toprule
    &   \shortstack{Contract\\to fiber} & \shortstack{Converges to\\fiber distribution} & Fast setup & Fast sampling \\
    \midrule
    Conditional models & \uncertain & \cmark & \xmark & \cmark \\
    \hspace{0.4cm} + Fiber loss (ours) & \cmark & \cmark & \xmark & \cmark  \\
    Guidance (ours) &  \cmark & \uncertain & \cmark & \xmark \\
    \bottomrule
\end{tabular}

    }
    \label{tab:method-comparison}
\end{wraptable}
We show that we can create diverse, high-quality fiber samples on various subject models and datasets, making invariance auditing practical at scale. To our knowledge, this is the first time guidance of pretrained models has been exploited as a way to examine invariances. Together, as shown in \cref{tab:method-comparison}, these methods are complementary: Guided invariance auditing enables rapid iteration during model development, while conditional fiber models support large-scale statistical analysis.

In summary our contributions are:
\begin{itemize}
\item The fiber loss as a regularizer to control invariance levels in fiber model training.
\item A training-free invariance auditing method via guidance of pretrained generative models.
\item A controlled benchmark with precisely known fiber distributions, comparing a comprehensive range of fiber model designs.
\item Large-scale invariance auditing of popular vision models on natural and medical images, confirming known and revealing previously unreported failure modes.
\end{itemize}


\section{Related Work}

Existing work can be broadly categorized into model inversion techniques, that construct an image from a given representation and generative models which sample from the fiber. 

\noindent
\textbf{Model inversion techniques.} The work of \citet{erhan2009visualizing} introduces gradient ascent in the input space to maximize the activations of given specific neurons. Following this work, several attempts have been made to invert a given set of activations (i.e. intermediate representations) using various techniques, including gradient descent \citep{mahendran2015understanding}, (de-)convolutional networks \citep{zeiler2014visualizing, dosovitskiy2016inverting} and GANs \citep{dosovitskiy2016generating}. Output matching model inversion attacks apply these concepts to reconstruct input data information from model outputs \citep{yang2025deep}. Several works have attempted to curate sets of natural adversarial images for certain domains in a model-agnostic way \citep{hendrycks2021natural, rahmanzadehgervi2024vision, mayer20256}, but the size and domain coverage of these sets remains limited. More recently, \citet{sammani2023visualizing} fit a local linear model, similar to LIME \citep{ribeiro2016should}, to measure the level of sensitivity to changes in each pixel of the image. Model inversion seeks to invert a subject model deterministically, while the goal of invariance auditing is to efficiently explore a diverse, representative set of samples from the fiber.

\noindent
\textbf{Invariance auditing based on generative models.} The pioneering work of \citet{rombach2020making} first formulated the task of sampling from fibers as a conditional generative problem, and utilized coupling-based normalizing flows to draw such samples. \citet{eulig2024reconstructing} utilize this approach to analyze the invariances of CT image denoisers. \citet{bordes2021high} replaced the generative model by a conditional diffusion model, which leads to better generative quality. They then compare the invariances of self-supervised models and thoroughly examine properties of the backbone and projector embeddings. The works of \citet{appalaraju2020towards, ericsson2021well, zhao2020makes} explore different approaches by utilizing deep image priors \citep{ulyanov2018deep}. As shown in \cite{bordes2021high} however, this leads to low quality samples. While these works lay the foundation for invariance auditing, it has not been widely adopted as a tool for trustworthy model design and diagnostics. We argue that this is largely due to the initial cost of training a conditional generative model for every subject model. We fill this gap by presenting an efficient method for invariance auditing using pretrained, unconditional diffusion models.

\noindent
\textbf{Learning and controlling invariances.} A complementary line of work focuses on explicitly encouraging or discouraging models to rely on certain invariances during training (e.g. \citet{arjovsky2019invariant}). An extended discussion of this topic can be found in \cref{app:IRM}. Invariance auditing serves a complementary visualization and diagnostic purpose: It identifies undesirable or lacking invariance structure in given subject models. Invariance control can then be leveraged to improve the models, and repeated invariance auditing will verify success. In \cref{sec:IRM} we illustrate this workflow on a dataset from \citet{arjovsky2019invariant}. Invariance auditing is used to make reliance of a baseline model on a known spurious correlation in the data apparent. Using invariance control to eliminate this reliance, invariance auditing confirms that the resulting classifier ignores the spurious correlation.

\section{Invariance Auditing and Fibers}
\label{sec:Fibers}

Given a subject model $\phi$ and a representation created by that model $\phi(x) = h$ the goal of invariance auditing is to sample from the fiber (or invariant set)
\begin{align}
    \phi^{-1}(h) := \{x: \phi(x) = h \}.
\end{align}
This task is often further narrowed down to conditional generative modeling with the target distribution  
\begin{align}
    & p(x|h) \propto p(x) \delta(\phi(x) - h)
\end{align}
The underlying distribution $p(x)$ can be freely chosen and does not have to be the same as the dataset distribution that the subject model was trained on. For example in a safety relevant context, $p(x)$ may be chosen to explicitly contain many high risk datapoints. Using a conditional generative model, one can inspect samples from the learned density $p_\theta(x|h)$, which intuitively answer the question:
\begin{highlightbox}
    \textit{Which inputs have the same representation and are thus indistinguishable for the subject model?}
\end{highlightbox}
Due to limitations in numerical accuracy and convergence it is generally impossible to sample $\tilde{x}$ that achieves exactly $\phi(\tilde{x}) = h$ when $h$ is continuous.
Likewise, we generally cannot rely on having access to multiple samples from any single fiber in our dataset. Based on the above characterization of the fiber, we define two separate goals of an accurate fiber model: \textbf{Fidelity} and \textbf{Consistency}. 

\noindent
\textbf{Fidelity} describes the ability of the fiber model to produce samples close to the fiber. Following \citet{bordes2021high}, we evaluate the fidelity of fiber models using the discrepancy between the target representation and the representation of fiber samples, which we call \textit{fiber loss}
\begin{align}
    \mathcal{L}_\text{fiber} = \text{dist}(\phi(x), \phi(\tilde{x})).
    \label{eq:FiberLoss}
\end{align}
One could choose any metric to measure the difference in representations, for the remainder of this paper, we use the Euclidean metric $|| \phi(x) - \phi(\tilde{x})||_2^2$. In \cref{app:MetricChoice} we show that this is a robust choice, which generally implies low fiber loss in other metrics as well. The fiber loss can also be used as an efficient objective for training and guidance of fiber models, which was not explored in prior work. In \cref{sec:ColorMNIST} we will show that it is necessary to include this additional loss term to achieve higher levels of fidelity. The meaning of the fiber loss depends on the scale of the representations. For evaluating fidelity we provide the fiber losses achieved with nearest neighbors from the dataset and by augmenting the original image as a point of comparison. Achieving fiber losses lower than the nearest neighbor means we can gain insights from fiber samples about the subject model that we could not have gained from the dataset alone.

\noindent
\textbf{Consistency} describes the ability of the fiber model to capture the desired distribution on the fiber $p(x|h)$. As described earlier, one generally has neither access to samples, nor likelihood evaluation of this density. Instead, we estimate consistency by sampling from the learned marginal density
\begin{align}
    p_\theta(x) = \int p_\theta\big(x | h = \phi(x')\big)\, p(x')\,dx'
\end{align}
We can compare $p_\theta(x)$ and $p(x)$ via various sample-based metrics, e.g. FID. In \cref{sec:ColorMNIST} we construct an example, where the $p(x|h)$ is explicitly known, which allows us to benchmark architectures and training archetypes in terms of their KL-divergence on the fiber. 

We emphasize that neither fidelity nor consistency can be falsely improved by a bias in the fiber model, which is important: A bias in the fiber model can result in fiber samples that have the wrong distribution on the fiber -- or do not lie on the fiber at all. Fidelity and consistency verify that the obtained fiber samples are correct, or expose the bias in the fiber model if present.

\subsection{Invariance Auditing using Conditional Generative Models}
\label{sec:CondFiberModel}

\begin{wrapfigure}{r}{0.5\textwidth}
    \centering
    \includegraphics[width=\linewidth]{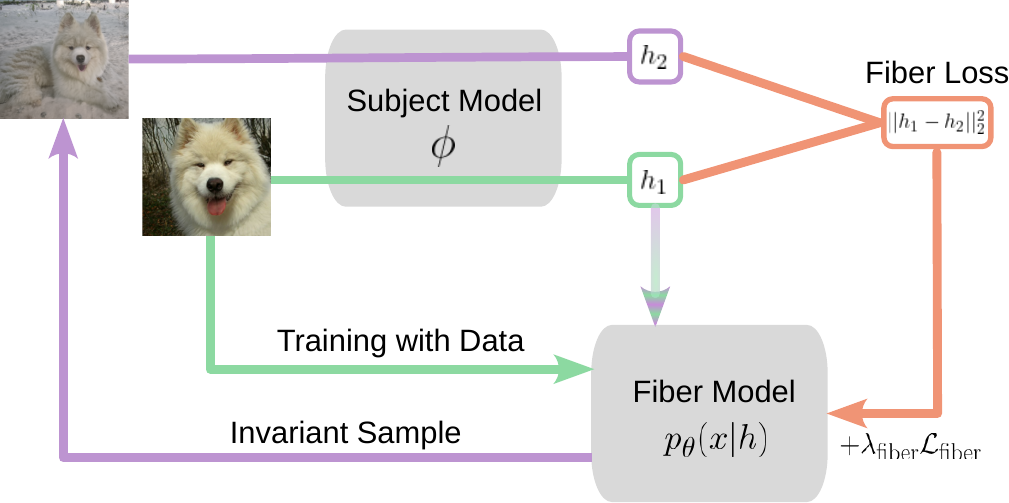}
    \caption{Conditional invariance auditing: The fiber model is trained to generate images conditioned on subject model representations $h_1$, producing samples with the same representation. The fiber loss verifies this and supplements training.}
    \label{fig:FrameworkOverviewConditional}
\end{wrapfigure}
Starting with \citet{rombach2020making}, invariance auditing has been formulated as a conditional generative modeling problem.  Different generative paradigms have been used for invariance auditing, yet their relative performance in terms of fidelity and consistency has not been systematically evaluated. 
In \cref{sec:ColorMNIST} we comprehensively benchmark different models for invariance auditing on a modified version of color MNIST \citep{rombach2020making}, where the fiber distribution is explicitly known. 

In this work we propose to use the fiber loss (\cref{eq:FiberLoss}) during training for improved fidelity; \cref{fig:FrameworkOverviewConditional} gives an overview of the workflow. Generative models with a fast sampling process can incorporate it at little additional cost, and as we show in \cref{sec:ColorMNIST}, doing so produces samples significantly closer to the target fiber. We also compare full-dimensional models to injective models, which are specifically crafted for lower-dimensional manifolds.
The intrinsic dimension of the fiber may differ from that of $x$. More precisely, assuming regularity of the subject model $\dim x - \text{rank}(D\phi) = \dim \phi^{-1}(h)$.
We show that injective models offer improvements especially at high regularization strength (see \cref{sec:ColorMNIST}).



\subsection{Invariance Auditing using Unconditional Diffusion Models}
\label{sec:UncondFiberModel}

One major drawback of training conditional generative models for invariance auditing is that for every subject model a new fiber model has to be trained from scratch. Depending on the task, training a conditional fiber model can even be more costly than training the subject model itself. Therefore, it is desirable to enable the use of large, pretrained generative models to create fiber samples, which are not conditioned on the subject model representation. We achieve this by utilizing guidance, specifically non-linear diffusion trajectory matching (NDTM) \citep{pandey2025variational}. 
The goal of NDTM is to guide the sampling process of a pretrained diffusion model to minimize some terminal cost function $\mathcal{L}_\text{terminal}(x_0)$ at the final sampling timestep $t=0$. In our case, we set $\mathcal{L}_\text{terminal}$ as the fiber loss of \cref{eq:FiberLoss} in order to only obtain samples from the subject model fiber. Since this loss can only be computed at time $t=0$, NDTM approximates 
\begin{align}
    & \mathbb{E}[\mathcal{L}_\text{terminal}(x_0, h)] \approx \mathcal{L}_\text{terminal}(\mathbb{E}[x_0|x_t], h) = ||\phi(\mathbb{E}[x_0|x_t]) - h||_2^2
\end{align}
via Tweedie's formula in a simulation-free manner. To inject guidance of strength $\gamma_t$ into the sampling process, a correction to the noisy sample at each timestep $u_t$ is made, where the guided diffusion process alternates between a denoising step and a sample correction step
\begin{wrapfigure}{r}{0.55\linewidth}
    \centering
    \includegraphics[width=\linewidth]{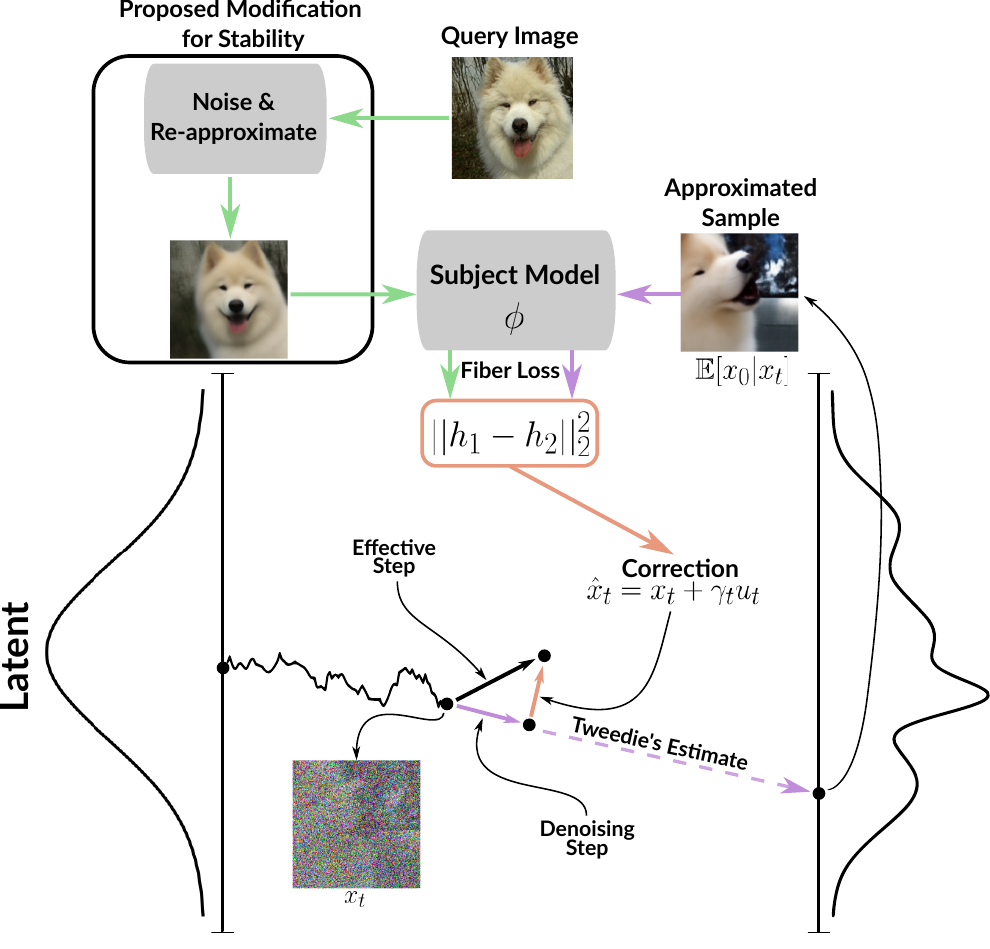}
    \caption{Guided invariance auditing: At each denoising step, Tweedie's estimate extrapolates the sample to the final step; the fiber loss against the target representation then corrects the sample. No finetuning or training is required.}
    \label{fig:FrameworkOverviewUnconditional}
\end{wrapfigure}
\begin{align}
    & \hat{x}_t = x_t + \gamma_t u_t \\
    & x_{t-1} = \text{DenoisingStep}(\hat{x}_t).
\end{align}
In every sample correction step, $u_t$ is obtained by performing $N$ optimization steps via gradient descent to minimize the following loss function:
\begin{align}
    \mathcal{L} = \mathcal{L}_\text{terminal} + \kappa_t \mathcal{L}_\text{control} + \tau_t \mathcal{L}_\text{score}
\end{align}
$\mathcal{L}_\text{control}$ and $\mathcal{L}_\text{score}$ are both regularizers that penalize the magnitude of $u_t$ and the deviation from the unguided score respectively.
Using this technique we can create samples from the fibers of a subject model using a pretrained, unconditional diffusion model, without the need to perform any finetuning. This is, to our knowledge, the first time pretrained model guidance has been proposed for analyzing fibers. We provide additional details about regularizers and hyperparameter choices in \cref{app:CoarseNDTM}. 

For our experiments on ImageNet, we notice that the generation process can sometimes be unstable, leading to oversaturated and overtexturized images. The features $h$ are potentially computed from detailed, local structures of $x$ in the data space. However, Tweedie's estimate $\phi(\mathbb{E}[x_0|x_t])$ only reconstructs coarse structures at larger $t$. Assuming access to $x$, we instead propose the terminal loss
\begin{align}
    & \mathcal{L}_\text{terminal}(x_t, x'_t) = \left|\left| \phi(\mathbb{E}[x_0|x_t]) - \phi(\mathbb{E}[x_0|x'_t]) \right| \right|_2^2.
    \label{eq:LterminalNew}
\end{align}
Where, $x'_t \sim q(x'_t|x_0 = x)$ is the noising process of the query image to time $t$. Effectively, we noise and apply Tweedie's estimate to $x$ to obtain an approximation with the same level of detail as the estimate based on $x_t$. We find this provides $u_t$ with a training signal that is more aligned with the semantic content of the current generation stage. In \cref{app:CoarseNDTM} we compare samples obtained with our modified algorithm and standard NDTM. We present an algorithm of NDTM for fiber sampling with the definition of the regularizers and our modification in \cref{alg:NDTM} and show an overview in \cref{fig:FrameworkOverviewUnconditional}.

\begin{algorithm}[htb]
\caption{Fiber Sampling with NDTM Guidance}
\label{alg:NDTM}
\begin{algorithmic}[1]
\STATE \textbf{Input:} Pretrained diffusion model, Subject model $\phi$, Query image $x$, Guidance strengths $\gamma_t$, Regularizer weights $\kappa_t,\tau_t$, Learning rates $\rho_t$, Number of gradient steps $N$
\STATE \textbf{Output:} Fiber sample $\tilde{x}$ with $\phi(\tilde{x}) \approx \phi(x)$

\STATE $h \gets \phi(x)$, \quad $x_T \sim \mathcal{N}(0,\mathbf{I})$
\FOR{$t = T, \dots, 1$}
    \STATE $h_t \gets h$ \COMMENT{Modification: $x'_t \sim q(x'_t|x_0=x)$, \; $h_t \gets \phi(\mathbb{E}[x_0|x'_t])$}
    \STATE $u_t \gets 0$
    \STATE $\hat{x}_t \gets x_t + \gamma_t u_t$
    \FOR{$i = 0$ to $N-1$}
        \STATE $\mathcal{L}_\text{terminal} \gets \|\phi(\mathbb{E}[x_0|\hat{x}_t]) - h_t\|_2^2$
        \STATE $\mathcal{L}_\text{control} \gets \|u_t\|_2^2$
        \STATE $\mathcal{L}_\text{score} \gets \|\nabla \log p(\hat{x}_t) - \nabla \log p(x_t)\|_2^2$
        \STATE $\mathcal{L} \gets  \mathcal{L}_\text{terminal} + \kappa_t \mathcal{L}_\text{control} + \tau_t \mathcal{L}_\text{score}$
        \STATE $u_t \gets u_t - \rho_t \nabla_{u_t} \mathcal{L}$
        \STATE $\hat{x}_t \gets x_t + \gamma_t u_t$
    \ENDFOR
    \STATE $x_{t-1} \gets \mathrm{DenoisingStep}(\hat{x}_t)$
\ENDFOR
\STATE $\tilde{x} \gets x_0$
\STATE \textbf{return} $\tilde{x}$
\end{algorithmic}
\end{algorithm}

\section{Experiments}

\subsection{Color MNIST Benchmark}
\label{sec:ColorMNIST}
To benchmark different methods under known invariances, we construct a setup where the fiber and its distribution are known.
Based on the Color MNIST dataset from \citet{rombach2020making} (not to be confused with the like-named dataset from \citet{arjovsky2019invariant}, which we use in \cref{sec:IRM}), we randomly colorize the background with a distribution $p(c_0)$ and derive a corresponding foreground color $c_1$ deterministically from it, see Appendix \ref{sec:app_cmnist} for details. 
The subject model turns the color images back into grayscale, and a proper fiber model must learn the distribution of background colors as well as their deterministic relationship to  foreground colors.
We can thus measure both quality criteria independently: 
High fidelity requires preservation of digit shapes and the correct foreground/background relationship, which can be measured by fiber loss.
Consistency is high when the known $p(c_0)$ is reproduced, which can be measured by KL-divergence.
\begin{wrapfigure}{l}{0.65\linewidth}
    \centering
    \includegraphics[width=\linewidth]{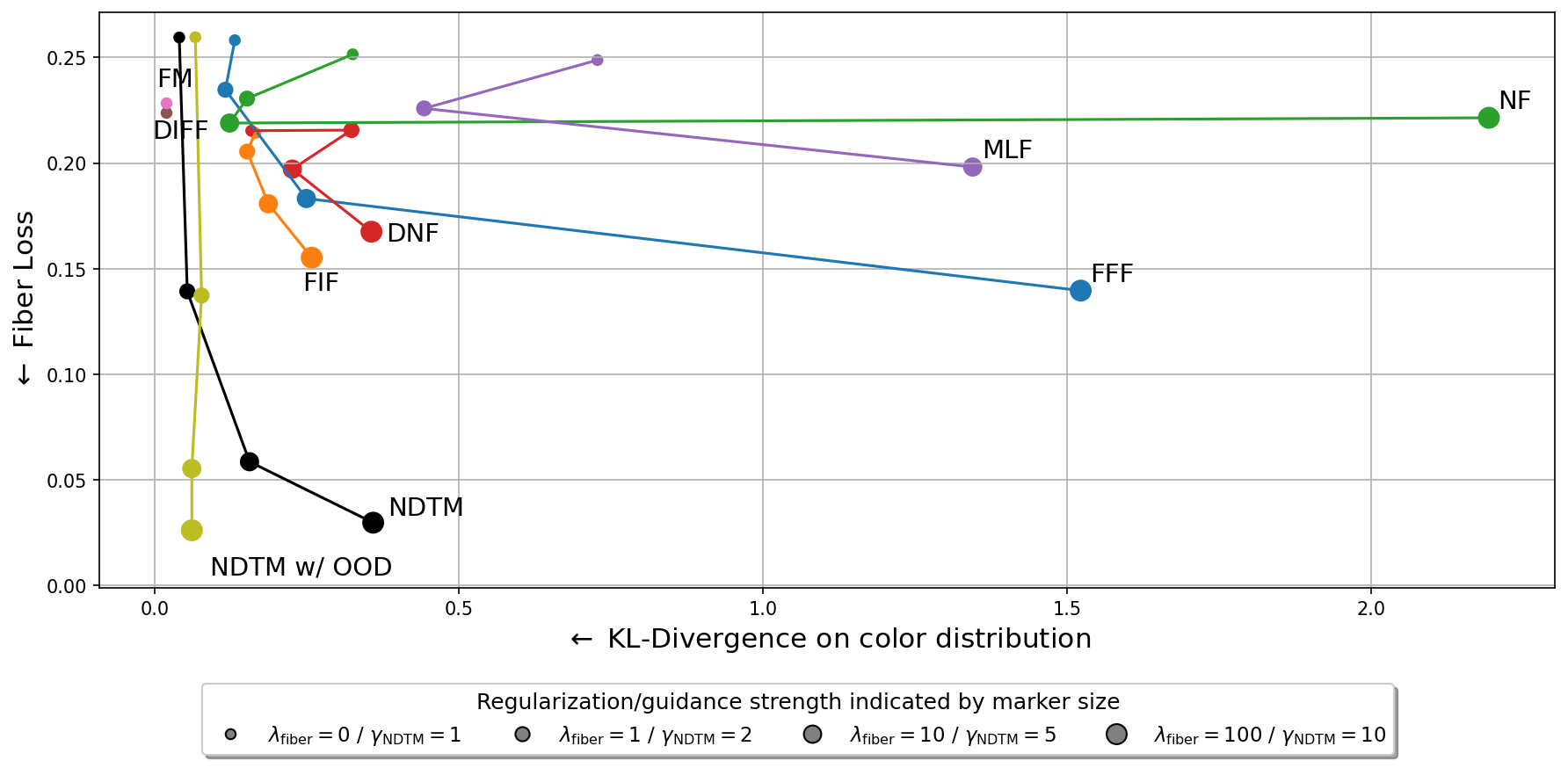}
    \caption{Benchmark of different invariance auditing methods and architectures on color MNIST. Fiber samples are evaluated by KL-divergence of background color distribution and the fiber loss (lower is better for both). Each colored line represents one method, with the marker size representing the strength of the training regularization $\lambda_\text{fiber}$ or guidance $\gamma_\text{NDTM}$.
    }
    \label{fig:ColorMNISTPareto}
\end{wrapfigure}
Within this setup, we perform a comparison between the design choices listed in \cref{sec:Fibers}. We train various classes of conditional fiber models and their injective variants where available. Architectures, as well as additional details and metrics are listed in \cref{sec:app_cmnist}. For NDTM we additionally examine an OOD setting where the correlation between fore- and background color is removed from the diffusion model (i.e. fiber model) training data. Even though the background color distribution during training is $p(c_0)$ for both diffusion models, NDTM w/ OOD may also adjust $c_0$ during guidance to yield the invariant color combination between fore- and background.


In \cref{fig:ColorMNISTPareto} we show the results of these experiments in terms of fiber loss vs. KL-divergence of the color distribution. Regarding the different generative paradigms we see significant differences in KL-divergence, with diffusion/flow matching models clearly coming out on top. Without regularization conditional fiber models all achieve similar values of fiber loss. At their baseline, (most) injective models perform marginally better in terms of fiber loss, and are more robust to high regularization parameters $\lambda_\text{fiber}$. Trained both with and without the correct relationship between colors, NDTM can achieve much better fidelity. The consistency compared to conditional diffusion models is only slightly worse with the right settings for the in-distribution case, the consistency of NDTM w/ OOD is not affected by the guidance strength showing that even a diffusion model pretrained on a broader distribution can serve for guided fiber sampling. Overall, diffusion-based conditional models achieve the best consistency, while NDTM attains substantially higher fidelity with minimal loss in consistency in the right settings.

\begin{figure}[htb]
    \centering
    \begin{subfigure}[b]{0.48\linewidth}
    \begin{tikzpicture}
        \node (img) {\includegraphics[width=\linewidth]{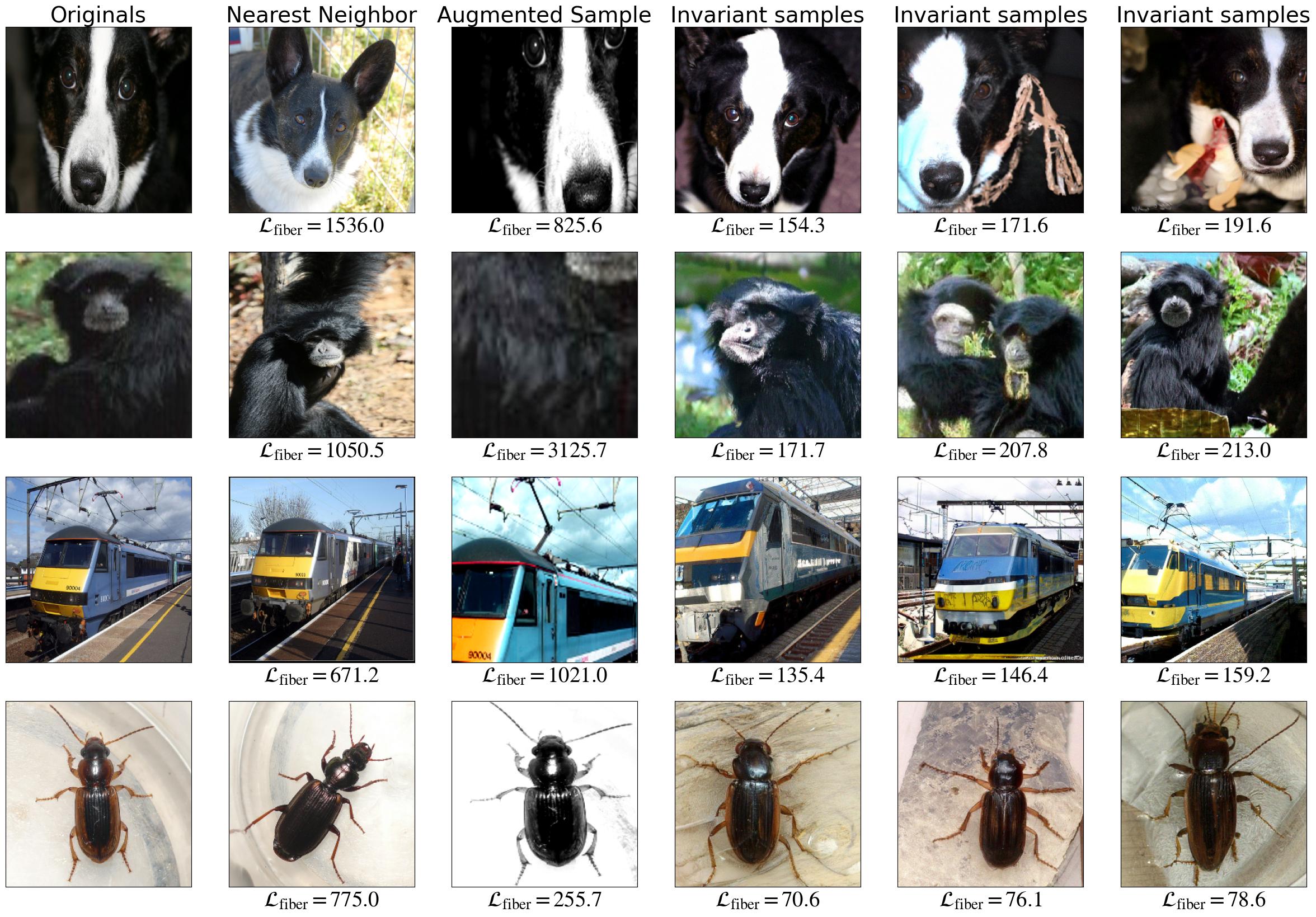}};
        \draw[black, very thick] 
            ($(img.south west)!0.5!(img.south east)$) --
            ($(img.north west)!0.5!(img.north east)$);
    \end{tikzpicture}
    \caption{Fibers of DINOv2}
    \label{fig:RandomImagenetSamples}
    \end{subfigure}
    \hfill
    \begin{subfigure}[b]{0.48\linewidth}
        \begin{tikzpicture}
        \node (img) {\includegraphics[width=\linewidth]{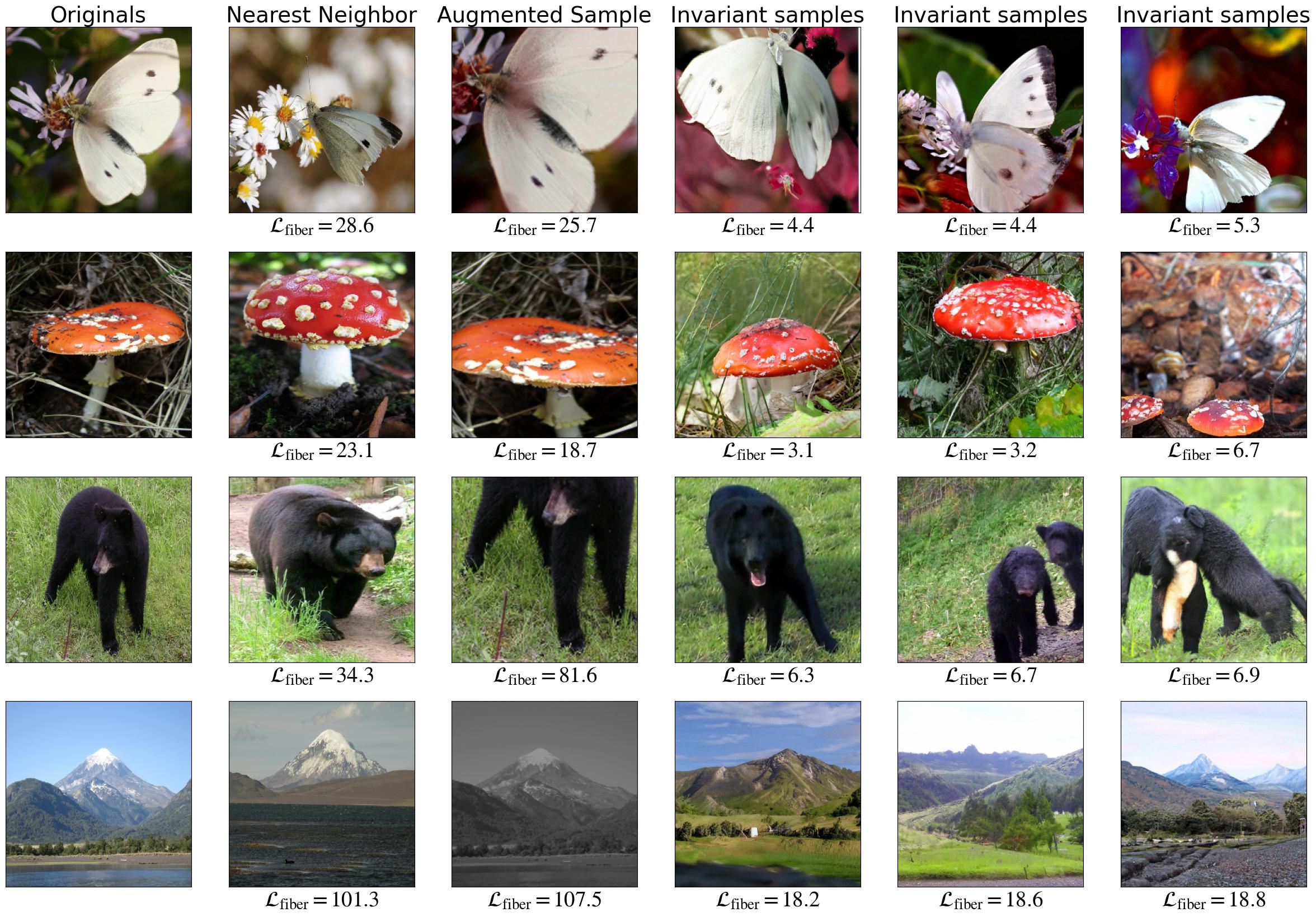}};
        \draw[black, very thick] 
            ($(img.south west)!0.5!(img.south east)$) --
            ($(img.north west)!0.5!(img.north east)$);
    \end{tikzpicture}
    \caption{Fibers of InceptionV3}
    \label{fig:RandomInceptionSamples}
    \end{subfigure}
    \caption{Fiber samples drawn with NDTM on ImageNet. The first three columns show the original, nearest neighbor and the original with DINOv2 training augmentations. Best viewed zoomed in.}
\end{figure}

\vspace{-0.4cm}

\subsection{Invariances on ImageNet}

We showcase the scalability of NDTM by analyzing the invariances of pretrained models on ImageNet. We follow a similar setup to \citet{bordes2021high} using a diffusion model trained on ImageNet from \citet{dhariwal2021diffusion} to create samples that are invariant under the representations of DINOv2 ViT-B/14 backbone. Crucially, using guidance allows us to perform this sampling with a pretrained diffusion model, without any fine-tuning. In the fiber samples in \cref{fig:RandomImagenetSamples} we observe that image orientation and main object position are also preserved, confirming the finding of \citet{bordes2021high} that SSL-trained model backbones are not necessarily invariant to the augmentations used for training. In comparison to nearest neighbors from the dataset and originals with DINOv2 training augmentations our fiber samples achieve high fidelity. We provide fiber losses of all subject models and large image datasets in \cref{tab:fiber-losses}, which confirms that NDTM can contract to fibers even for high-dimensional, complex representations. 
\begin{wraptable}{r}{0.5\textwidth}
    \centering
    \caption{FIDs and top-5 accuracy of a pretrained ConvNeXt-L on fibers of DINOv2 and InceptionV3. We measure FID both with a TensorFlow (TF) and PyTorch (PT) Inception model. Baselines show the unconditional diffusion FID and classifier performance on ImageNet.}
    \resizebox{\linewidth}{!}{


\begin{tabular}{cc|ccc}
    \toprule
    Subject Model & Fiber Model & TF FID & PT FID & \shortstack{Classifier\\top-5 accuracy}  \\
    \midrule
    \multicolumn{2}{c|}{Baseline on ImageNet} & 26.21 & - & $95.62\%$ \\
    DINOv2 & NDTM & 28.34 & 29.68 & $81.24\% \pm 1.29$ \\
    InceptionV3 & NDTM & 33.52 & 3.81 & $51.48\% \pm 1.89$ \\
    \bottomrule
\end{tabular}
    }
    \label{tab:MetricsImagenet}
\end{wraptable}
The reported FIDs in \cref{tab:MetricsImagenet} confirm that NDTM generated samples still follow the ImageNet distribution to similar quality as the unconditional base model. Note that our goal is \textit{not to improve} the quality of generated images, but inspect the invariances of the subject model with the help of a good, pretrained generative model. These results confirm that NDTM enables large-scale, training-free inspection of invariances while preserving generative quality.

Next we inspect InceptionV3 \citep{szegedy2016rethinking} to ask the question: What changes in images are not captured by FID and could lead to low values despite distribution shift? To this end, we sample from the fiber of the PyTorch pretrained InceptionV3 model and compare FIDs computed both with this model in PyTorch and the TensorFlow InceptionV3 model that was originally used in \citet{heusel2017gans} in \cref{tab:MetricsImagenet}. When computing the FIDs for the DINOv2 fiber, both networks yield slightly different, but similar FID values. In contrast, sampling from fibers of the Inception network itself reveals that FID can be artificially low when errors of the generator align with the fibers of the metric network, exposing the blind spots of such representation-dependent evaluation. Inspecting these blind spots with the help of fiber samples in \cref{fig:RandomInceptionSamples} and the top-5 
accuracies of a pretrained classifier applied to the fiber samples shows that members of the InceptionV3 fiber still have high visual quality, but potentially confuse different classes.

\subsection{Invariances on Cue Conflict Dataset}
\label{sec:CueConflict}

We turn our attention towards a case where the target representation is created by an out-of-distribution input. We compare the representations of DinoV2 ViT-B/14 backbone and those after the final pooling layer of a ResNet50 model, pretrained on ImageNet, on the cue conflict dataset from \cite{geirhos2018imagenet}, which stylizes images with different textures to decouple them from object shape.
\vspace{0.3cm}
\columnratio{0.45, 0.55}
\globalcounter{figure}
\begin{paracol}{2}

\begin{leftcolumn}
\begin{figure}
    \includegraphics[width=\linewidth]{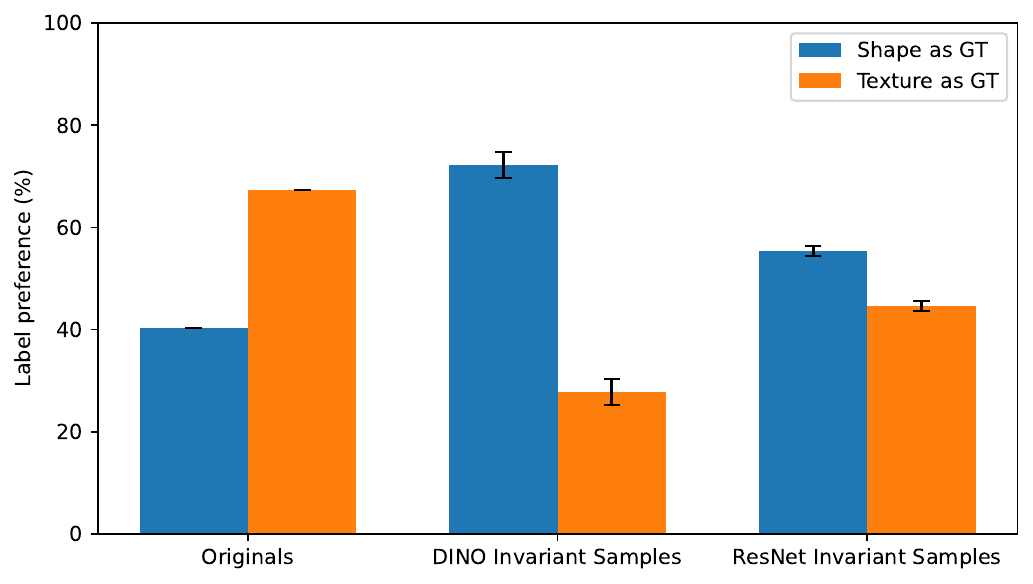}
    \caption{Preference for texture (orange) and shape (blue) labels of ConvNeXt-L on original data and subject model fibers. Cases where none of the two labels were predicted or both labels coincide are filtered out.}
    \label{fig:CueConflictBias}
\end{figure}
\end{leftcolumn}

\begin{rightcolumn}
\begin{figure}
    \begin{tikzpicture}
        \node (img) {\includegraphics[width=\linewidth]{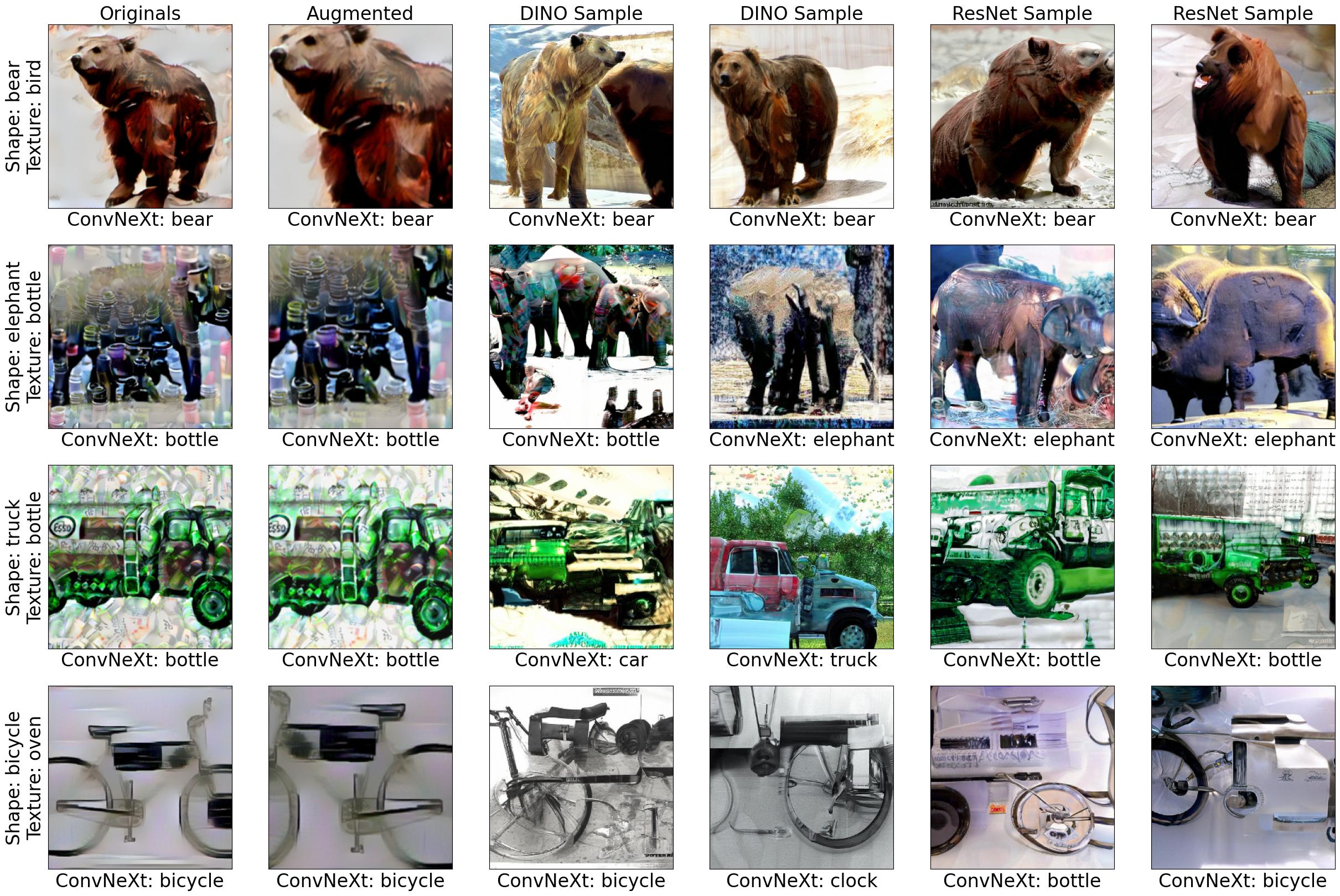}};
        \draw[black, very thick] 
            ($(img.south west)!0.355!(img.south east)$) --
            ($(img.north west)!0.355!(img.north east)$);
        \draw[black, very thick] 
            ($(img.south west)!0.675!(img.south east)$) --
            ($(img.north west)!0.675!(img.north east)$);
    \end{tikzpicture}
    \caption{Fiber samples of both subject models on the cue conflict dataset. Below each image we show the label predicted by a ConvNeXt-L. All models are trained on ImageNet. Best viewed zoomed in.}
    \label{fig:RandomCueConflictSamples}
\end{figure}
\end{rightcolumn}

\switchcolumn
\vspace{-0.7cm}

We use the same ImageNet diffusion model from the previous section for NDTM to obtain the fiber samples shown in \cref{fig:RandomCueConflictSamples}. Under each image we show the label predicted

\end{paracol}

\vspace{-0.55cm}

by an ImageNet pretrained ConvNeXt-L \citep{liu2022convnet}. Although in this OOD setting fiber losses are higher than on ImageNet, they remain well below nearest neighbors. We can see that samples from both invariant sets differ significantly from ImageNet, indicating that the subject models are not invariant to the distribution shift introduced by texture stylization. Inspecting the fibers more closely we find that DINOv2 representations generally preserve global shape information, while ResNet representations capture local shape information, which is in line with \citet{burgert2025imagenet}. We quantify these biases in terms of the label preference used in \citet{geirhos2018imagenet}. 

The ConvNeXt-L classifier is applied to originals and fiber samples to compare the accuracy when using the shape and texture labels of each image respectively. Specifically, in cases where the classifier agrees with exactly one of the two labels we report the percentage of times the shape or texture label was chosen respectively as \textit{label preference} in \cref{fig:CueConflictBias}. As reported by \cite{geirhos2018imagenet}, ImageNet-trained convolutional models like ConvNeXt and ResNet are also biased towards texture, but we find that this bias is not yet expressed in the final pooling layer of ResNet-50. This indicates that the shape information is not lost in the fiber, rather it is not being chosen as the basis for classification by the last, fully-connected layer. In contrast, we find that DINOv2 representations exhibit a bias towards encoding image content by its shape. In fact, while the classifier recovers the shape label 25\% of the time on its own, applied to DINOv2 fibers this increases to 35\%.

\subsection{Combining Invariance Auditing with Invariance Learning and Control}
\label{sec:IRM}

We demonstrate how invariance auditing and invariance control can work together to first expose and then eliminate a model's reliance on spurious features. We inspect two subject models on an MNIST variant from \citet{arjovsky2019invariant}, which we call "causal MNIST". 
The task is to predict whether a colored MNIST digit is high ($\geq 5$) or low ($< 5$). Color is strongly correlated with the label during training, 
\begin{wrapfigure}{r}{0.55\textwidth}
    \centering
    \includegraphics[width=\linewidth]{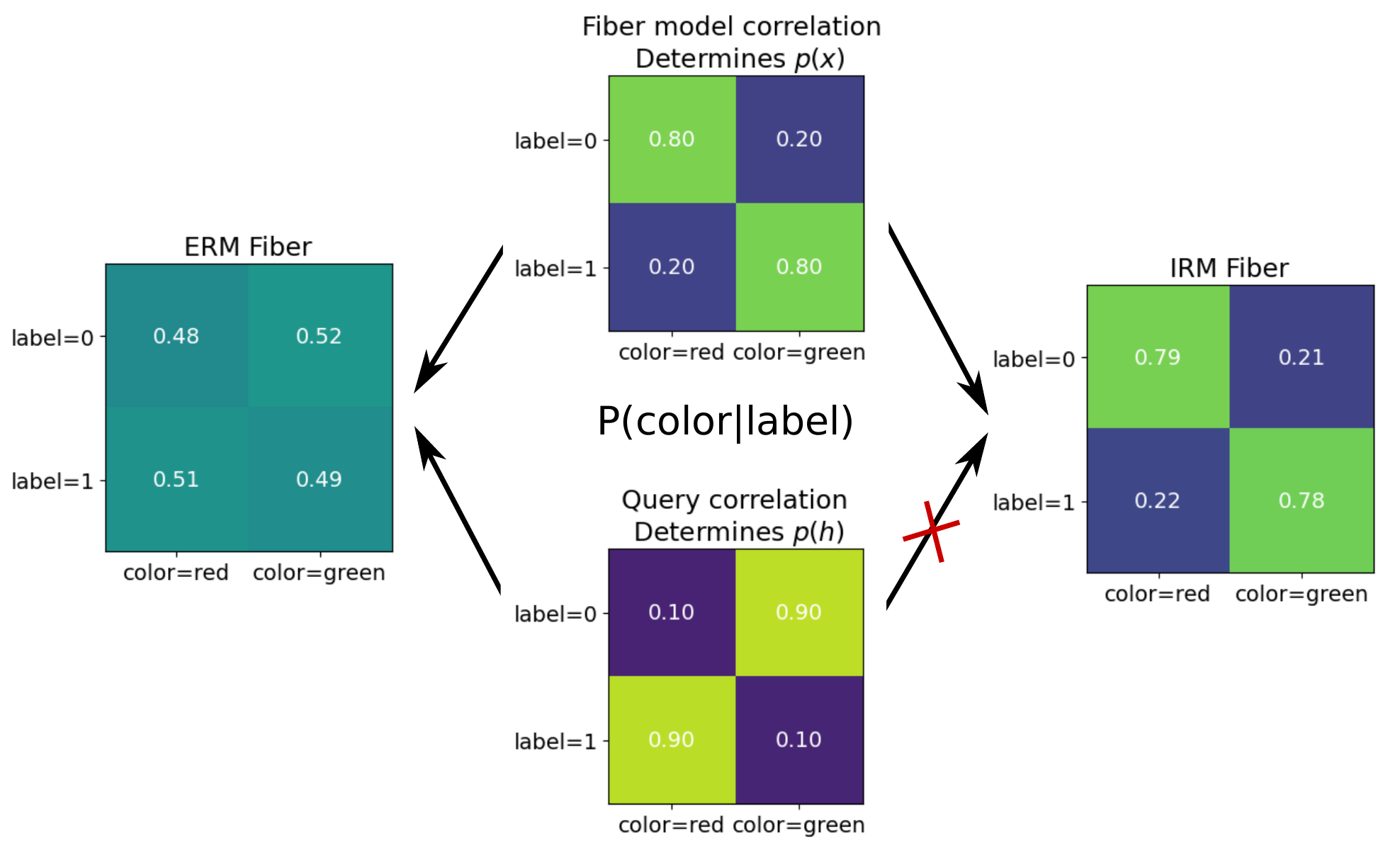}
    \caption{Color--label correlation in fiber samples on causal MNIST. The fiber model is trained with 80\% color--label correlation, determining $p(x)$. Query representations are obtained from inputs with reversed correlation, determining $p(h)$. A shift in correlation on the fiber away from $p(x)$ indicates that the subject model prediction is not invariant to color.}
    \label{fig:DomainShift}
\end{wrapfigure}
but at test time this correlation is flipped; models that rely on this predictor therefore do not generalize. We train the fiber model for guidance with the training correlation between label and color, which determines $p(x)$. The query images are obtained using the reversed color correlation, which determines $p(h)$. If color affects the representation $h$ the fiber samples $p(x|h)$ should have a different color correlation than $p(x)$. This is what invariance auditing detects in a naively trained subject model (ERM), which indicates that it takes the color shortcut. We then retrain the subject model using a method designed to control invariances (IRM), which removes this false reliance on a spurious correlation. Invariance auditing then confirms that the resulting model correctly uses only the digit shape to predict the label (see \cref{fig:DomainShift}). Full details are given in \cref{sec:IRMvsERM}.

\subsection{Invariances on CheXpert}
\label{sec:Chexpert}
\begin{wrapfigure}{r}{0.6\textwidth}
    \vspace{-1.5cm}
    \centering
    \includegraphics[width=\linewidth]{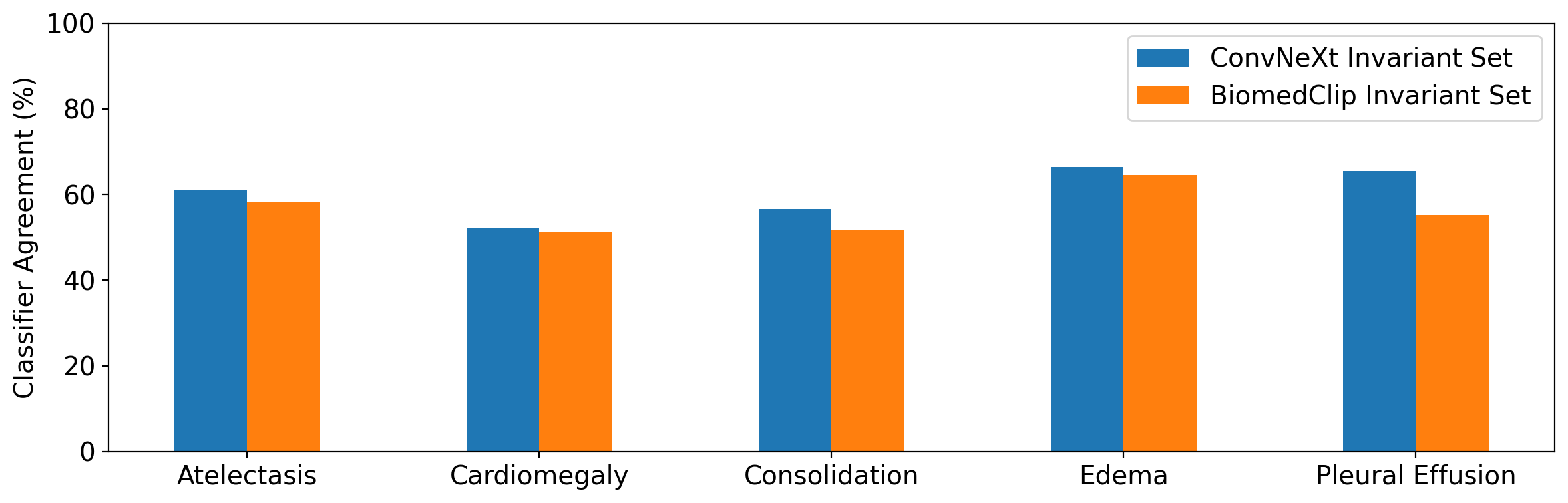}
    \caption{Cross-model agreement on CheXpert invariances. For images both models classify correctly, we measure how often they assign the same label to each other’s fiber samples.}
    \label{fig:ChexpertAgreement}
\end{wrapfigure}

To inspect fibers in medical image analysis, we train two classifiers of different quality on five classes of the CheXpert dataset \citep{irvin2019chexpert}. For the first classifier we freeze the BiomedClip foundation model \citep{zhang2023biomedclip} and train a classification head. The second classifier is a ConvNeXt trained end-to-end. Averaged across the five classes the classifiers achieve an AUROC of 0.73 and 0.85 respectively; further details in \cref{app:Chexpert}. The fiber loss is computed at the classifier logits, which yields images with the same class probabilities. We measure classifier agreement by drawing fiber samples for originals where both classifiers are correct, then labeling them with the other model. 

\begin{wrapfigure}{l}{0.58\textwidth}
    \centering
    \includegraphics[width=0.48\linewidth]{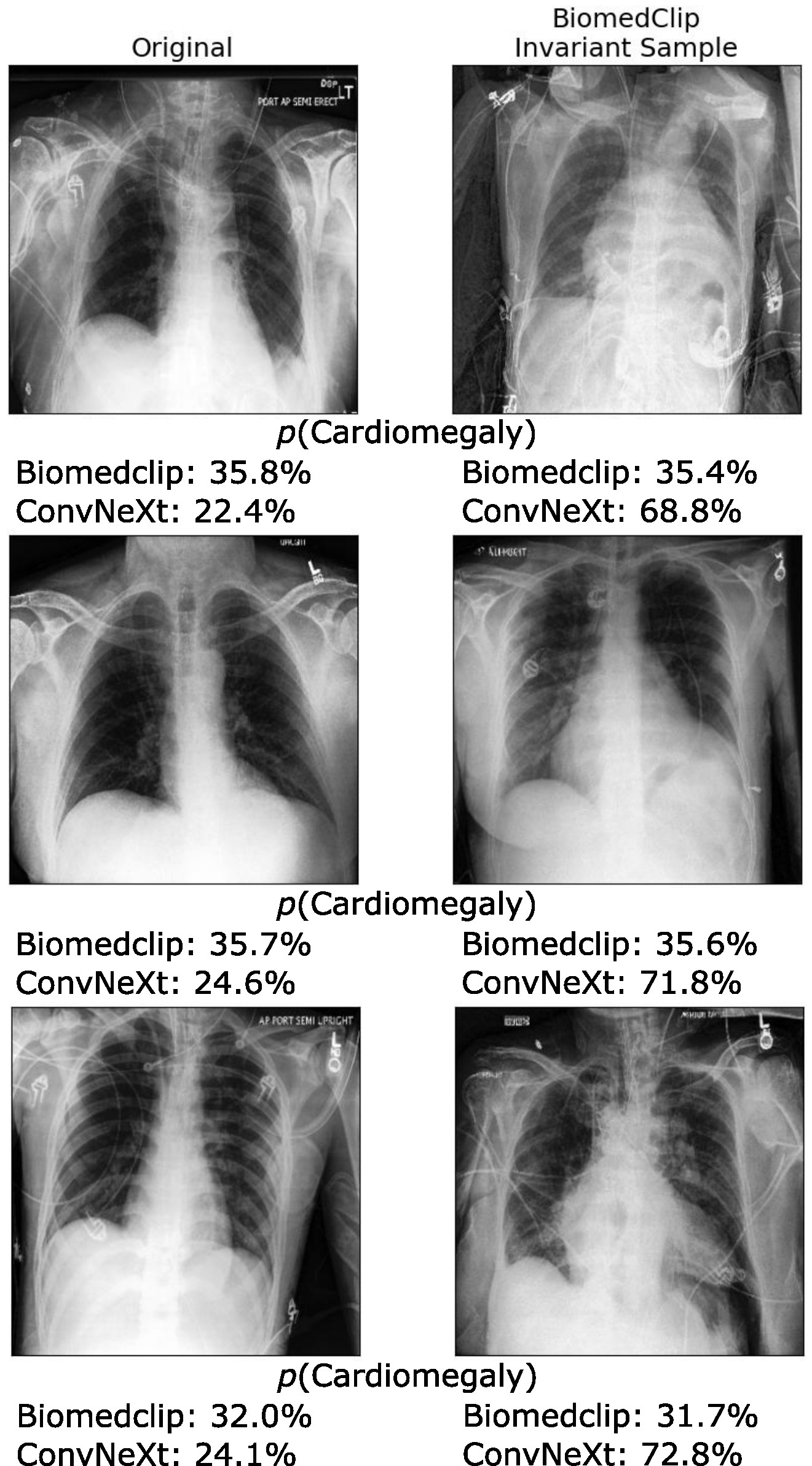}
    \unskip\ \vrule\
    \includegraphics[width=0.48\linewidth]{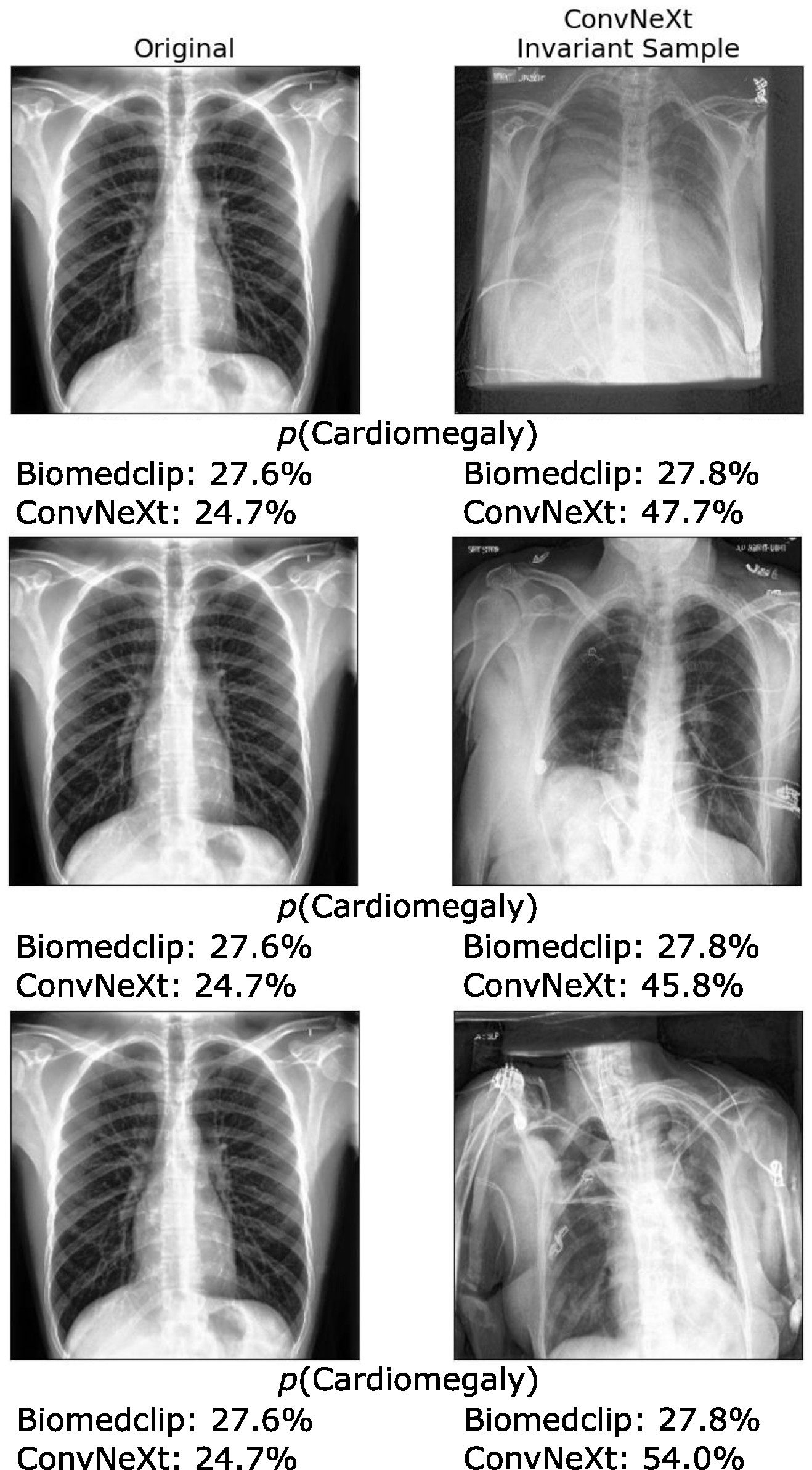}    
    \caption{Originals with negative cardiomegaly diagnosis and samples of the fiber from BiomedClip (left) and ConvNeXt (right), obtained by our search for contradicting samples.}
    \label{fig:ChexpertExamples}
\end{wrapfigure}

We plot the rate of agreement as the percentage of fiber samples where both classifiers stay with the original label in \cref{fig:ChexpertAgreement}. As we can see, models disagree on their fibers indicating that at least one of the models has not learned desirable fibers. In \cref{fig:ChexpertExamples} we show samples from the fiber, where the ground truth label for cardiomegaly on the original was negative, but the other classifier now predicts a positive label. We find that the ConvNeXt classifier detects critical changes in the BiomedClip fiber, while the BiomedClip classifier mostly flags images with unclear changes.

In \cref{app:Chexpert} we repeat these experiments for two identically trained end-to-end classifiers. This shows how fiber sampling enables targeted stress-testing of representations in safety-critical settings. In particular, the observed disagreement between classifiers highlights that invariances can be model-dependent and that undesirable blind spots may only become apparent when viewed from an external reference. This suggests a practical workflow for invariance auditing: training a complementary reference model, or leveraging a pretrained one provides an additional signal to detect semantic changes in fiber samples and to identify invariances that are undesirable for the task at hand. 

In the recently introduced natural medical adversarial images \citep{mayer20256} the authors show that, when prompting vision-language models (VLMs) like Qwen-2B with images that show rare medical cases (e.g. anatomy), they will give answers that align with typical anatomy, disregarding the information in the images. Our goal is to find out whether this decision is made by the language model or the information is already lost at the vision model embedding.

\clearpage

\begin{wrapfigure}{r}{0.6\textwidth}
    \includegraphics[width=\linewidth]{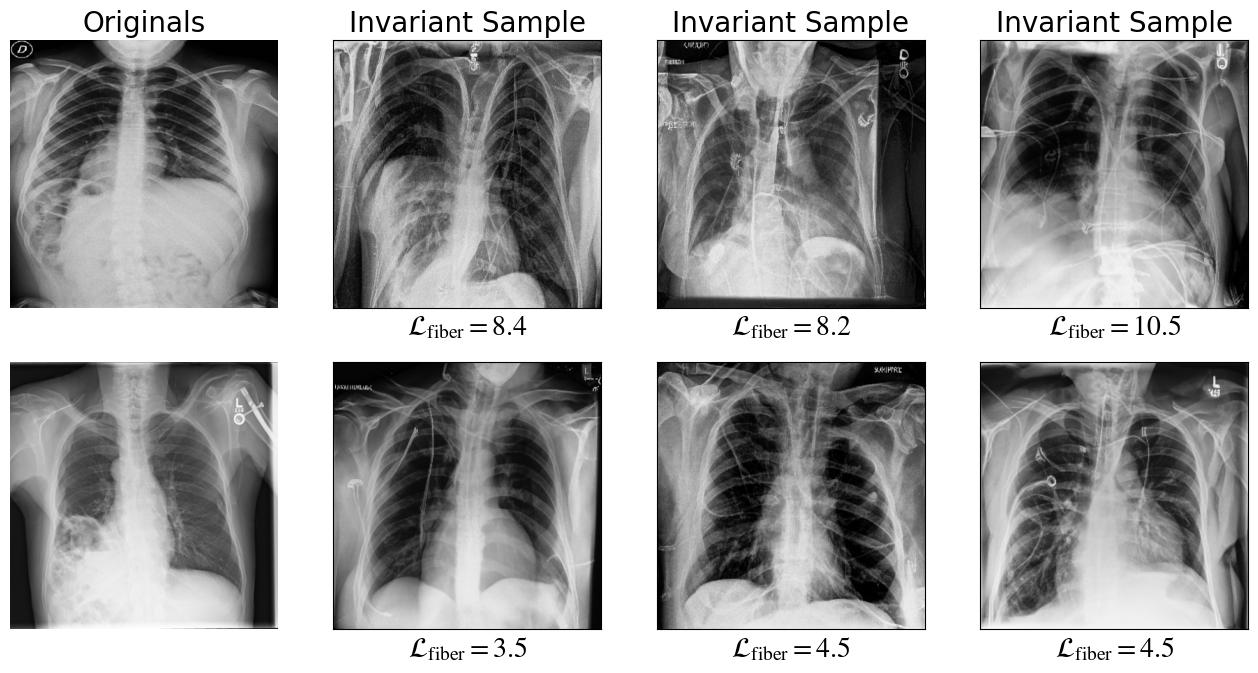}
    \caption{Invariances of the Qwen-2B vision model on a "situs inversus" image (shown on the left). The atypical anatomy does not appear in the selected samples.}
    \label{fig:QwenExamples}
    \vspace{-0.4cm}
\end{wrapfigure}

To this end, we create fiber samples of "situs inversus" cases from \citet{mayer20256} where the heart is located on the right side of the body (from the patient's perspective). Our subject model is the Qwen-2B pretrained vision encoder. Fiber samples in \cref{fig:QwenExamples} suggest that positional information of the heart is weakly encoded by the vision encoder. This explains the finding of \citet{mayer20256}, that even tuning the prompt to take into account rare medical occurrences, the VLM was unable to give the correct response. Beyond this, the experiment illustrates the flexibility of invariance auditing: it enables the targeted analysis of rare cases that would otherwise be overlooked by population-level evaluation. Without requiring large-scale datasets to detect systematic failures, we verify for all four situs inversus images that typical anatomy occurs in the fiber -- a failure mode that aggregate metrics would be unlikely to surface.

\section{Conclusion}
\label{sec:Conclusion}

In this work we presented two complementary improvements for invariance auditing. On the one hand, the contraction of conditional generative models to the fiber is improved using our fiber loss, increasing their reliability. On the other hand NDTM-based guidance produces invariant samples without any model training. Using this framework, we show that we can rediscover and visualize several findings from literature, including texture bias of CNNs and anatomical blind spots of VLMs. 
There remains a trade-off between spending compute on training a conditional fiber model and optimizing the cost function during inference when using guidance, making the two methods complementary. This naturally suggests a usage pattern -- using NDTM for rapid iteration during model development and conditional fiber models for large-scale statistical analysis. Combining both is a promising future direction.

With this expedited process we demonstrate how invariance auditing can be integrated into the design and evaluation of trustworthy models. Our experiments illustrate its merits by revealing a wide range of different invariances. Invariance auditing can give more detailed insights into the sources of known biases, at the same time it can also expose unknown failure-cases in modern vision and self-supervised models, such as omission of clinically relevant features in classifier representations. As such, invariance auditing presents a unified solution for exposing model blind spots. While it does not by itself prescribe how to remove them, it provides the critical first step by making them visible and measurable -- a practical foundation for iterative development of more robust and trustworthy models.

\section*{Acknowledgments and Funding Transparency Statement}
We would like to thank Elias Eulig for support and fruitful discussions, as well as Leon Mayer, Piotr Kalinowski and Lena Maier-Hein for providing swift, early access to the medical adversarial examples. 
This work is supported by Deutsche Forschungsgemeinschaft (DFG, German Research Foundation) under Germany’s Excellence Strategy EXC-2181/1 - 390900948 (the Heidelberg STRUCTURES Cluster of Excellence).
It is also supported by the German Federal Ministry of Education and Research (BMBF) (project EMUNE/031L0293A).
AR acknowledges support by the Carl-Zeiss-Stiftung (Projekt P2021-02-001 "Model-based AI").
The authors acknowledge support by the state of Baden-Württemberg through bwHPC and the German Research Foundation (DFG) through grant INST 35/1597-1 FUGG.
We ran experiments on NVIDIA A100 GPUs of the bwForCluster Helix.
UK thanks the Klaus Tschira Stiftung for their support via the SIMPLAIX project.
None of the authors received any other additional revenues in relation to this work in the last 36 months.


\bibliography{references}
\bibliographystyle{plainnat}

\newpage
\appendix
\onecolumn

\section{Sample Refinement using Latent Gradient Descent}
\label{sec:SampleRefinement}
\begin{figure}[htb]
    \includegraphics[width=\linewidth]{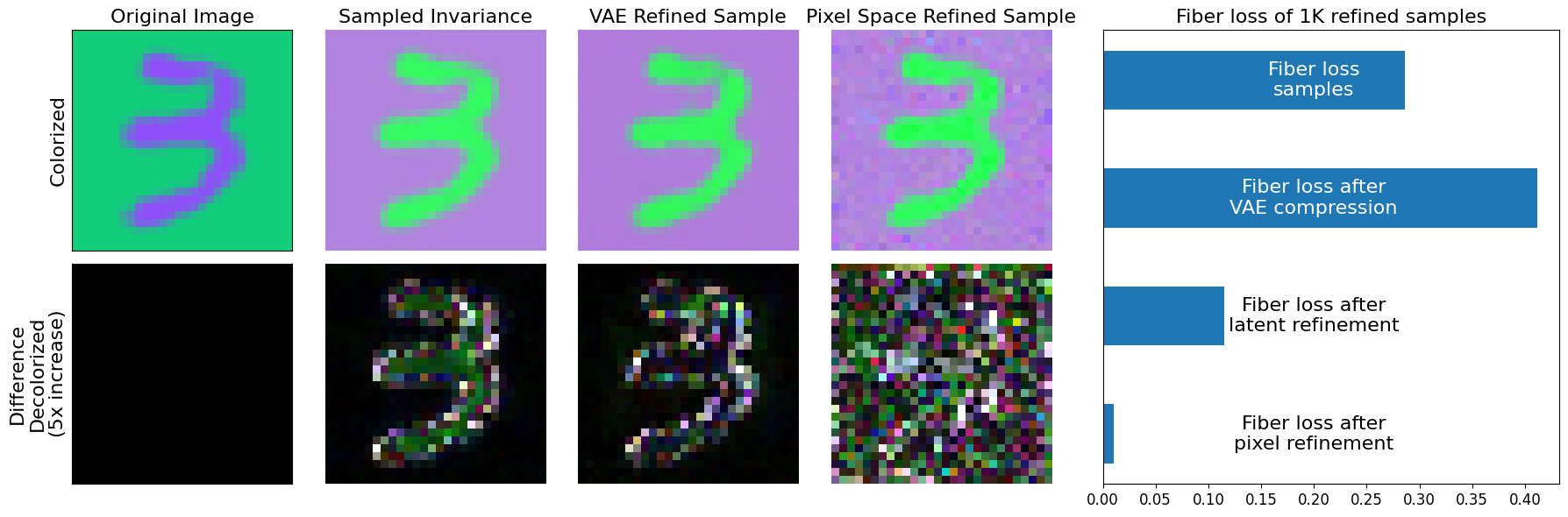}
    \caption{Sample refinement on a colorMNIST example. From left to right: Original image, invariant sample, sample refined with gradient descent in a VAE latent space and sample refined with gradient descent in pixel space. Below each image we show the difference of the decolorized image to the decolorized original. Note that even for non-zero difference the fiber loss may be small if they fall exactly in the invariances of the almost lossless autoencoder subject model, which is achieved by pixel level gradient descent. On the right we show the fiber loss improvement when applying sample refinement to fiber samples of a conditional normalizing flow fiber model on color MNIST.}
    \label{fig:SampleRefinementMNIST}
\end{figure}

Depending on the desired level of fidelity, the invariance sampling process may not contract sufficiently to the fiber, achieving a fiber loss that is low enough. We can further refine the sample to reduce the fiber loss by applying a few ($\sim 100$) gradient descent optimization steps directly on the sample, minimizing the fiber loss. Doing so directly in data space however, is likely to lead to pixel-level adjustments, which are not semantically meaningful, similar to adversarial attacks. We show this effect in \cref{fig:SampleRefinementMNIST}. For this reason, one needs to regularize the refinement process towards a natural image prior, for example \citet{mahendran2015understanding} introduce several regularizing terms to this end. We find that a good natural image prior is enforced by performing sample refinement in some semantic latent space, e.g. of an autoencoder, as shown in \cref{alg:SampleRefinement}. In \cref{fig:SampleRefinementMNIST} we show the improvements in fiber loss and resulting samples gained for by refining samples in a VAE latent space using gradient descent to minimize the fiber loss. The latter leads to a much smoother result, fixing the color correlation and keeping the digit shape. This is owed to a more natural landscape for optimization. Crucially, gradient descent in data space leads to manipulating single pixel values to arrive at an invariance under the VAE of the subject model (i.e. the VAE applied to the decolorized images). While this may be interesting in some applications, in this setting we are interested in semantic variation, which is not encoded in these changes.

\begin{algorithm}[htb]
\caption{Latent Sample Refinement using an Autoencoder}
\label{alg:SampleRefinement}
\begin{algorithmic}[1]
\STATE \textbf{Input:} Encoder $E$, Decoder $D$, Fiber sample $\tilde{x}$, Target $x$, Subject model $\phi$, Learning rate $\eta$, Number of gradient steps $N$
\STATE \textbf{Output:} Refined fiber sample $\hat{x}$

\STATE $\tilde{z}_0 \gets E(\tilde{x})$
\STATE $h \gets \phi(x)$
\FOR{$i = 0$ to $N-1$}
    \STATE $\tilde{x}_i \gets D(\tilde{z}_i)$
    \STATE $\tilde{h}_i \gets \phi(\tilde{x}_i)$
    \STATE $\mathcal{L}_i \gets \|\,h - \tilde{h}_i\|_2^2$
    \STATE $\tilde{z}_{i+1} \gets \tilde{z}_i - \eta\,\nabla_{\tilde{z}_i} \mathcal{L}_i$
\ENDFOR
\STATE $\hat{x} \gets D(\tilde{z}_N)$
\STATE \textbf{return} $\hat{x}$
\end{algorithmic}
\end{algorithm}

\section{Experimental Details}
\label{app:ExperimentalDetails}
Code to reproduce all of our experiments is available at \url{https://github.com/vislearn/InvarianceAuditing}.
\subsection{Colored MNIST}
\label{sec:app_cmnist}

\subsubsection{Dataset}

\begin{figure}[htb]
\centering
    \includegraphics[width=.4\linewidth]{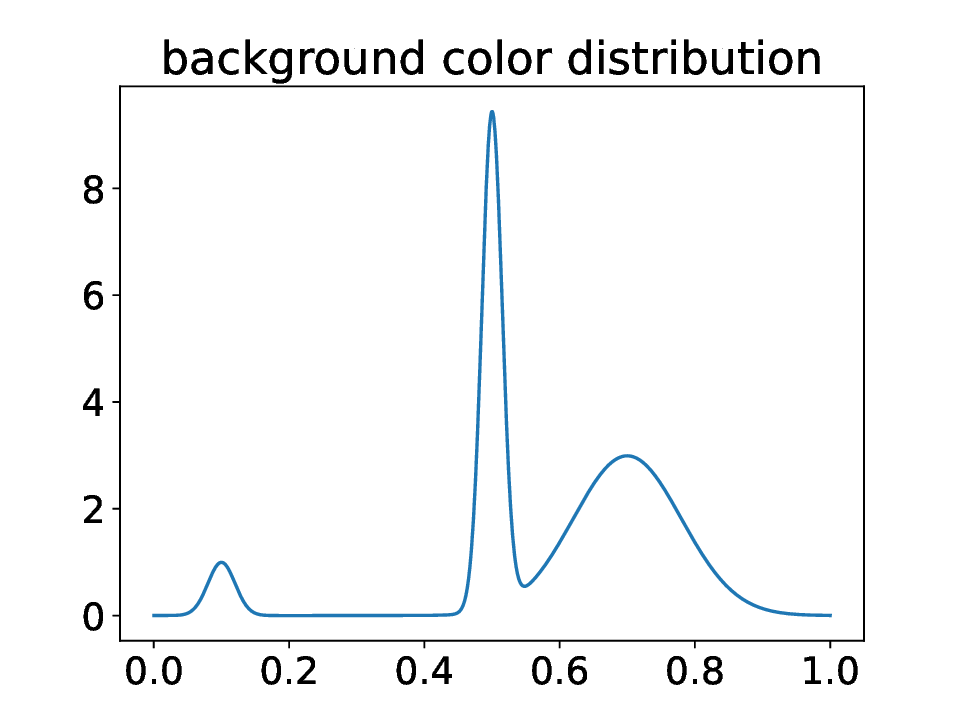}
    \includegraphics[width=.4\linewidth]{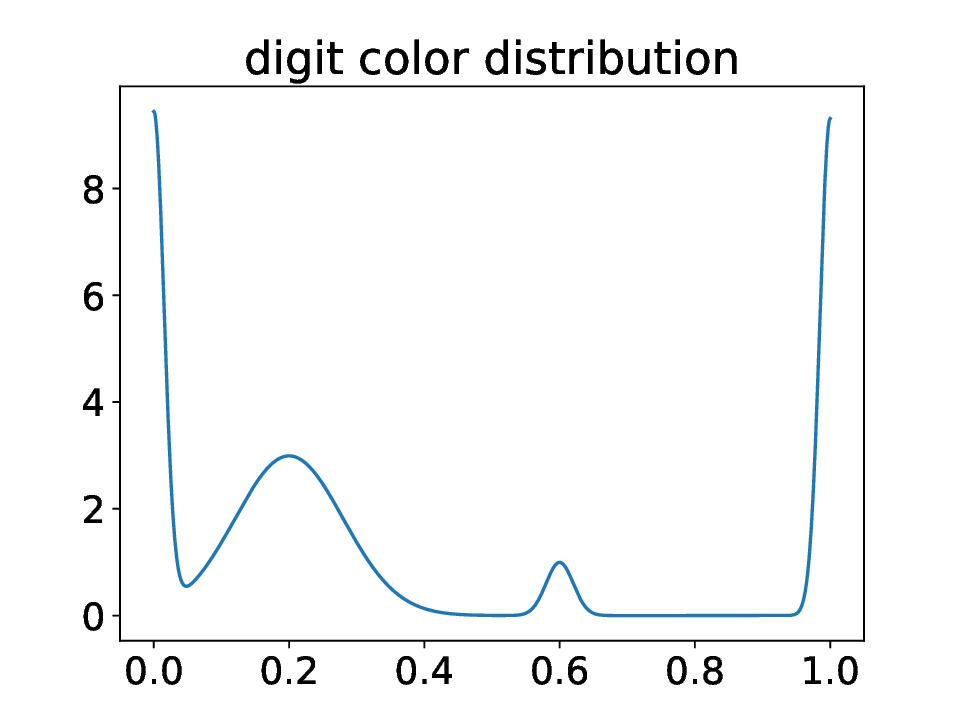}
    \caption{Gaussian mixture distributions for sampling the color values for each channel to colorize MNIST.}
    \label{fig:col_dist}
\end{figure}

The dataset is based on MNIST images, which we color according to a fixed relation between foreground and background color.
For each image, we obtain a random background color $c_0 \in \mathbb{R}^3$ and the corresponding foreground color $c_1 \in \mathbb{R}^3$. 
Each pixel is then colored according to its brightness.
For this we sample for every color channel a value $c^i$ i.i.d. from a Gaussian mixture distribution and compute the colored image by
\begin{equation}
    \vb{x}_{\text{colored}, i} = (1-\vb{x}) \cdot c^i + \vb{x}\cdot((c^i+0.5) \mod 1).
    \label{eq:x_colored}
\end{equation}
That is, the background of the image gets colored in $c_0 = (c^1, c^2, c^3)$ and the digit in $(c_0+0.5) \mod 1$. This ensures that the digit always has a certain contrast to the background. The Gaussian mixture distribution we use to sample each $c^i$ is given by
\begin{align}
    p(c^i) = 0.6 \cdot \mathcal{N}(c^i, 0.7, 0.08^2) + 0.35 \cdot \mathcal{N}(c^i, 0.5, 0.015^2) + 0.05 \cdot \mathcal{N}(c^i, 0.1, 0.02^2)
\end{align}
and shown in \cref{fig:col_dist} on the left. The resulting digit color distribution is shown on the right.
The decolorize function is now the analytical inverse of equation \ref{eq:x_colored} which is
\begin{equation}
    \vb{x}_i = \frac{\vb{x}_{\text{colored}, i}-c^i}{((c^i+0.5) \mod 1) - c^i}.
    \label{eq:x_dcolored}
\end{equation}
We see now that if the digit color is not exactly $(c^i+0.5) \mod 1$, the intensity of corresponding channel will be different than the original one. In order to examine the more common case where the features $\phi(x) = h$ are in vector shape, we encode the decolorized images by an effectively lossless autoencoder.

\subsubsection{Model training}

We train three different model classes as conditional fiber models: Normalizing flows (NF) \citep{kobyzev2020normalizing}, free-form flows (FFF) \citep{draxler2024free} and diffusion (DIFF) / flow-matching models (FM) \citep{ho2020denoising, lipmanflow}. To answer the question whether injective models are more suited to learn the low-dimensional fiber manifold, we train injective equivalents where available. We choose denoising normalizing flows (DNF) \citep{horvat2021denoising} and multilevel flows (MLF) \citep{kingma2018glow} for models based on invertible neural networks and free-form injective flows (FIF) \citep{sorrenson2023lifting} as the injective equivalent of free-form flows.
To stay accurate to the setting of \citet{rombach2020making} all conditional fiber models and the unconditional generative model for NDTM on color MNIST are latent generative models, learning the distribution of a VAE latent space. For clarity, we stress that this is not the lossless VAE serving as part of the subject model. The VAE is made up of 7.9 million parameters, distributed on encoder and decoder, each containing a convolutional part and a fully connected residual net. We list the hyperparameters in \cref{tab:vae_hparams}.
\begin{table}[htb]
    \centering
    \begin{tabular}{c|c}
        blocks per residual net & 3 \\
        layers per block        & 3 \\
        layer size              & 512 \\
        latent dimension        & 54 \\
        activation function     & SiLU \\
        epochs                  & 700 \\
        lr scheduler            & one-cycle \\
        max lr                  & 0.0002 \\
        reconstr. loss weight   & 2000 \\
    \end{tabular}
    \caption{Hyperparameters of the fiber model VAE used for the color MNIST experiments.}
    \label{tab:vae_hparams}
\end{table}

We trained all fiber models for up to 1000 epochs and checked for convergence in epochs 100 and 250, in which case we would apply early stopping. In general, we chose the architectures such that each fiber model contains around 10-15M parameters in total (including the VAE). Despite a fiber dimension of $\dim \phi^{-1}(h) = 3$, for all injective models except free-form injective flows we found a latent dimension of six to yield the best results. We use the onecycle learning rate scheduler and tune the learning rate from a range of $[2\times10^{-4}, 2\times10^{-3}]$.

        


\subsubsection{Additional Results}

\begin{figure}[htb]
    \centering
    \includegraphics[width=\linewidth]{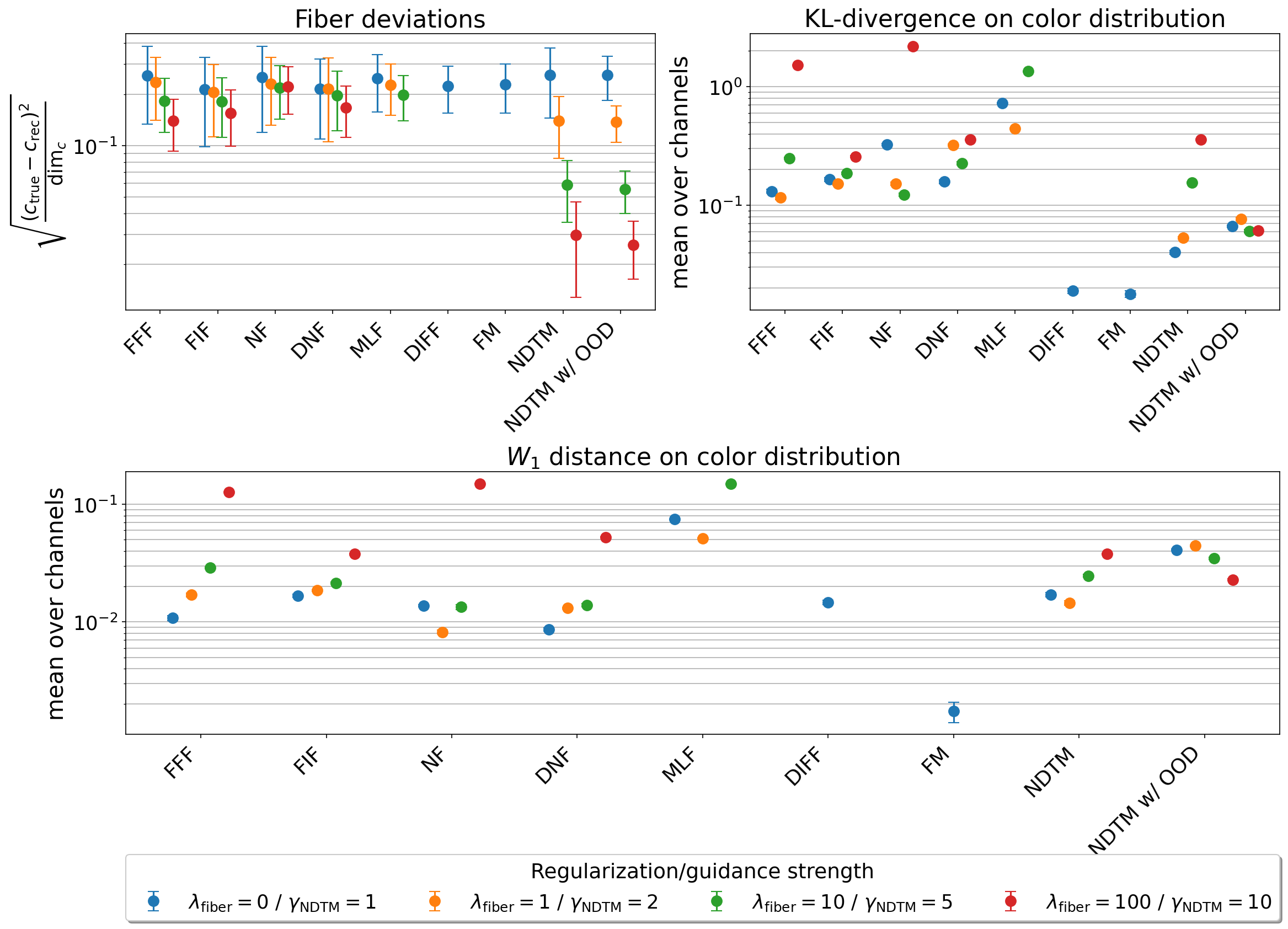}
    \caption{Comparative plot of fiber loss (top left), KL-divergence (top right) and $W_1$-distance (bottom) on the color distribution of the color MNIST benchmark. $\lambda$ gives the weight of the fiber loss during training of the conditional models, while $\gamma$ shows the guidance strength of NDTM. Standard deviations are taken across all sample sets and target embeddings (Note that for KL-divergence and $W_1$ distance the standard deviation is imperceptibly small for almost all models).}
    \label{fig:colorMNISTMetrics}
\end{figure}

In \cref{fig:colorMNISTMetrics} we provide comparisons of fiber losses, KL-divergences and Wasserstein distances on the color distribution for all models. These results show that there generally exists a trade-off between models that achieve good consistency and high fidelity. We also provide standard deviations for all metrics, obtained by creating ten (three for NDTM) independent samples for each fiber. Note that for the fiber loss, we also compute the standard deviation across different target embeddings, which is why it shows higher uncertainty. This encapsulates the difference in how well the methods can recover high fidelity samples across all possible target embeddings.

\subsection{ImageNet \& Cue Conflict}

In \cref{tab:Runtimes} we report estimates for the time required to train and sample with a conditional model like in \citet{bordes2021high} vs. sampling time of NDTM on a NVIDIA A40 GPU. We find that it only becomes faster to train a model when creating more than 9k samples, disregarding the time for hyperparameter optimization.

\begin{table}[htb]
    \centering
    \caption{Estimates of the time required to train and sample from a conditional fiber model vs. sampling via NDTM guidance on ImageNet. Numbers are based on sampling with 100 timesteps and $N=5$ optimization steps for NDTM on a NVIDIA A40 GPU.}
    \label{tab:Runtimes}
    \resizebox{0.5\linewidth}{!}{
    \begin{tabular}{ccc}
    \toprule
    \shortstack{Training\\conditional model} & \shortstack{Time per sample\\conditional model} & \shortstack{Time per sample\\NDTM} \\    
    \midrule
    $\sim 14000$ min & 0.3 min & 1.8 min \\
    \bottomrule
\end{tabular}
    }
\end{table}

In \cref{tab:fiber-losses} we list the fiber losses of experiments on ImageNet and the cue conflict dataset. This confirms that NDTM manages to contract closely to fibers in relation to e.g. nearest neighbors in a dataset, even when the target representations are created by OOD inputs.

\begin{table}[htb]
    \centering
    \caption{Fiber losses of all subject models on large image datasets. As a comparison we provide the fiber losses of the nearest neighbors from the dataset as well as the original image with typical DINOv2 training augmentations. Standard deviations are measured across 10 independently drawn samples per fiber.}
    \label{tab:fiber-losses}
    \resizebox{0.7\linewidth}{!}{
    \begin{tabular}{ccccc}
    \toprule
    \makecell{Subject Model} &
    \makecell{Dataset} &
    \makecell{$\mathcal{L}_{\text{fiber}}$ \\ invariant samples} &
    \makecell{$\mathcal{L}_{\text{fiber}}$ \\ nearest neighbor} &
    \makecell{$\mathcal{L}_{\text{fiber}}$ \\ augmented} \\
    \midrule
    DINOv2 & ImageNet & \textbf{273 $\pm$ 4} & 1225 & 440  \\
    Inception & ImageNet & \textbf{34.8 $\pm$ 0.7} & 123.9 & 94.9  \\
    DINOv2 & Cue Conflict & \textbf{487 $\pm$ 3} & 1069 & 946  \\
    ResNet & Cue Conflict & \textbf{38.1 $\pm$ 0.7} & 74.9 & 101.3 \\
    BiomedClip & CheXpert & \textbf{1.6 $\pm$ 0.0} & 23.3 & - \\
    ConvNext & CheXpert & \textbf{1.2 $\pm$ 0.0} & 10.9 & - \\
    Qwen & CheXpert & \textbf{5.5 $\pm$ 3.0} & 17.8 & - \\
    \bottomrule
\end{tabular}
    }
\end{table}

For evaluating ImageNet FID, fiber losses and classifier accuracy on the fiber in tables \ref{tab:MetricsImagenet} \& \ref{tab:fiber-losses}, we draw a set of 10k fiber samples.

\subsection{Causal MNIST}
\label{app:IRM}

In this section we illustrate how invariance auditing can be combined with methods that control which invariances a model learns. The resulting workflow can be used to develop models whose decisions are more robust and trustworthy. 

\subsubsection{Related work on controlling invariances}

Several works develop methods that encourage or discourage models to rely on certain invariances during training.
Invariant learning methods \citep{peters2016causal, arjovsky2019invariant, rosenfeld2021risks, creager2021environment, krueger2021out} aim to learn representations whose optimal predictors remain stable across multiple environments, thereby discouraging reliance on spurious correlations. Related approaches such as group distributionally robust optimization \citep{rahimian2019distributionally, oren2019distributionally, sagawa2019distributionally} promote invariances useful for generalization by penalizing worst-case risks. In the context of representation learning, methods based on data augmentations \citep{chen2020simple, oquab2024dinov2} implicitly enforce invariances to predefined transformations, while approaches like information bottleneck methods \citep{tishby2000information} discard nuisance factors, keeping only task relevant features. Conversely, several works study how to control and quantify sensitivity to specific attributes, for example through disentanglement objectives \citep{burgess2018understanding, locatello2019disentangling, wang2024disentangled}.

\subsubsection{Experimental Details}
\label{sec:IRMvsERM}

This section provides additional detail on the experiment performed in \cref{sec:IRM}. In this experiment we want to show how a combination of invariance auditing and invariance control can first reveal and then eliminate unwanted invariances. To this end we compare a model that is trained with classical empirical risk minimization (ERM) to one that is trained with invariant risk minimization (IRM) \citep{arjovsky2019invariant} on a version of color MNIST from the causal invariance literature. Note that this dataset is distinct from the one in \cref{sec:ColorMNIST}, which is based on the color MNIST dataset introduced by \citet{rombach2020making}, to distinguish them we will refer to the dataset from \citet{arjovsky2019invariant} as "causal MNIST". The binary classification task labels in causal MNIST are given by whether a digit is $<5$ or $\geq 5$. Noise is introduced to obtain the final labels by flipping them with a probability of $25\%$. Afterwards, the images are colored red or green with different probabilities, depending on three environments: In the two environments used for training the color ID coincides with the final label $80\%$ and $90\%$ of the time, while in the testing environment this correlation is flipped and it coincides with the final label only $10\%$ of the time. This way color serves as a stronger predictor than the digit itself, but the correlation is spurious and relying on color for prediction does not generalize to the test environment. In \citet{arjovsky2019invariant} it is shown that ERM relies on this spurious correlation and is not robust under domain shift, while IRM does not. All subject models are trained according to the publicly available implementation of \citet{arjovsky2019invariant}\footnote{\url{https://github.com/facebookresearch/invariantriskminimization}}.

Our aim is to exemplify a general workflow for trustworthy model design: We first show that invariance auditing reveals that ERM relies on the known spurious correlation. Adopting a method for invariance control like IRM then removes the undesired reliance on this predictor, making the model invariant to color. Invariance auditing then verifies that invariance control was successful. 

To perform invariance auditing we train a diffusion model on the environment with $80\%$ color -- label correlation for NDTM. This determines the color correlation of the underlying distribution $p(x)$ for invariance auditing. As target representations for invariance auditing we sample classifier logits $h \sim p(h)$ using input images from the environment with $10\%$ color -- label correlation. If no color information is present in $h$, i.e it is invariant to color, then the color -- label correlation in $p(x|h)$ should be the same as in $p(x)$. Otherwise we expect to see a bias towards the color -- label correlation of the inputs used to create $p(h)$ depending on how strongly the representation encodes color.

\begin{figure}[h]
    \centering
    \includegraphics[width=0.7\linewidth]{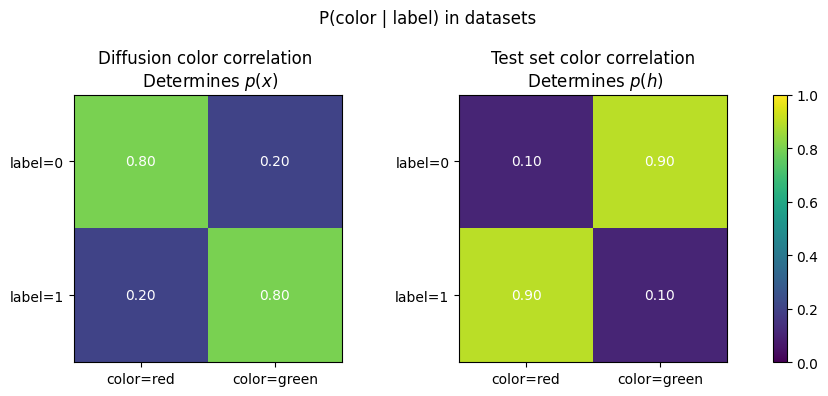}
    \includegraphics[width=0.7\linewidth]{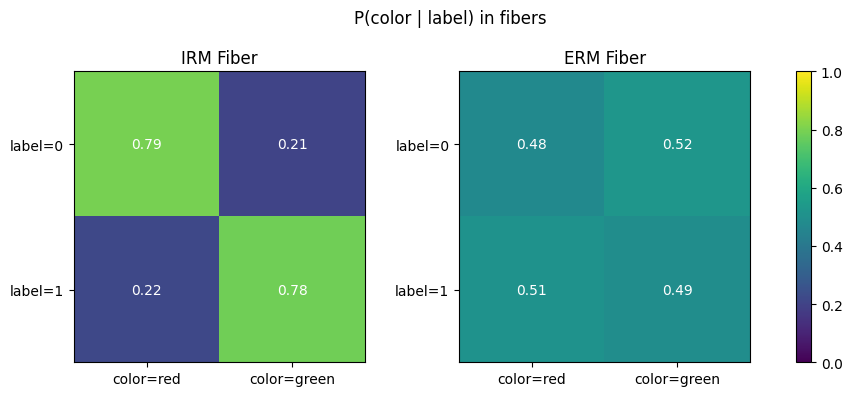}
    \caption{Top: Color -- label correlation in the diffusion model used for NDTM and the test data used to create the target representations for invariance auditing. Bottom: Color -- label correlation in the fiber samples of a model that does not encode color information in $h$ (IRM) and one that does (ERM).}
    \label{fig:ERMvsIRM}
\end{figure}

This is what we observe in \cref{fig:ERMvsIRM}. The robust IRM model exhibits no shift in color -- label correlation, while the ERM model does. This shows that when reliance on a spurious correlation is suspected, invariance auditing can be used to expose it. Pairing this diagnostic method with invariant learning methods like IRM can serve as a tool to first identify and then remove unwanted reliance on spurious features.

\subsection{Chexpert}
\label{app:Chexpert}

Our implementation of the subject model training is based on the Kaggle notebook available at \citet{chexpert_kaggle_notebook}. As our subject models we pick a ConvNeXt-tiny (end-to-end trained classifier) \cite{liu2022convnet} and a frozen BiomedClip feature extractor \citep{zhang2023biomedclip} with a 5-layer ResNet classification head. We train both classifiers for 36k steps on the CheXpert dataset \citep{irvin2019chexpert}, with an asymmetric loss function \citep{ridnik2021asymmetric}. As no test set is provided, we reserve a part of the original training set for validation, and use the original validation set for testing. ROC curves on test data are given in \cref{fig:ChexpertROCCurves}, upper panel. We calibrate the probabilities of each classifier via temperature scaling and set the classification threshold for each classifier, s.t. it is optimal under F1-score. The unconditional diffusion model architecture is a transformer UNet with 4 blocks and 2 layers each, totaling 18M parameters, which we trained for 120k steps. All models take a resolution of 384x384. 

\begin{figure}[htb]
    \centering
    \includegraphics[width=\linewidth]{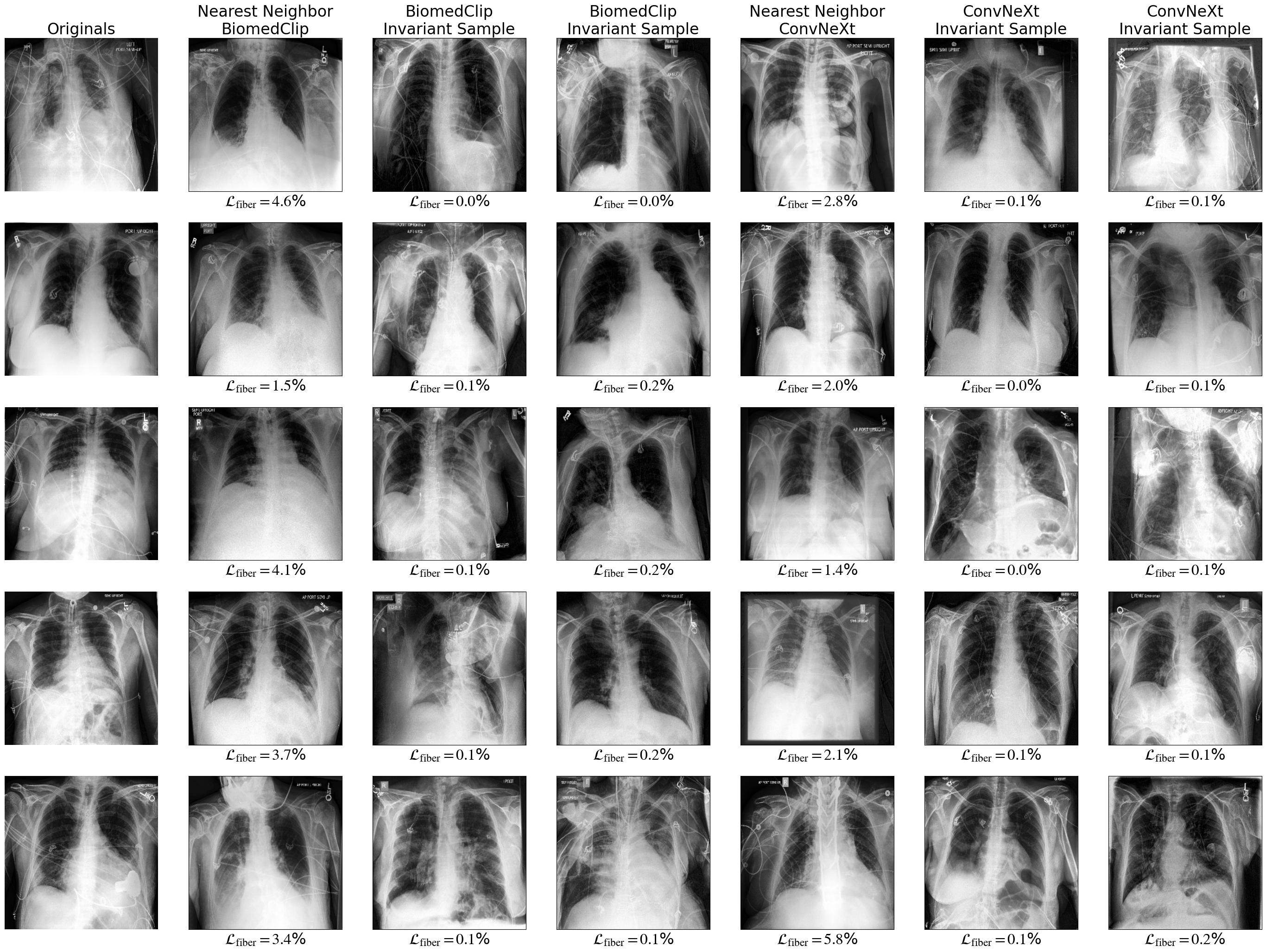}
    \caption{From left to right: Original image, nearest neighbor and two random invariant samples of BiomedClip, nearest neighbor and two random invariant samples from ConvNeXt. Below each image we give the summed absolute difference in predicted probabilities across all classes.}
    \label{fig:ChexpertRandomSamples}
\end{figure}

\begin{figure}[htb]
    \centering
    \includegraphics[width=\linewidth]{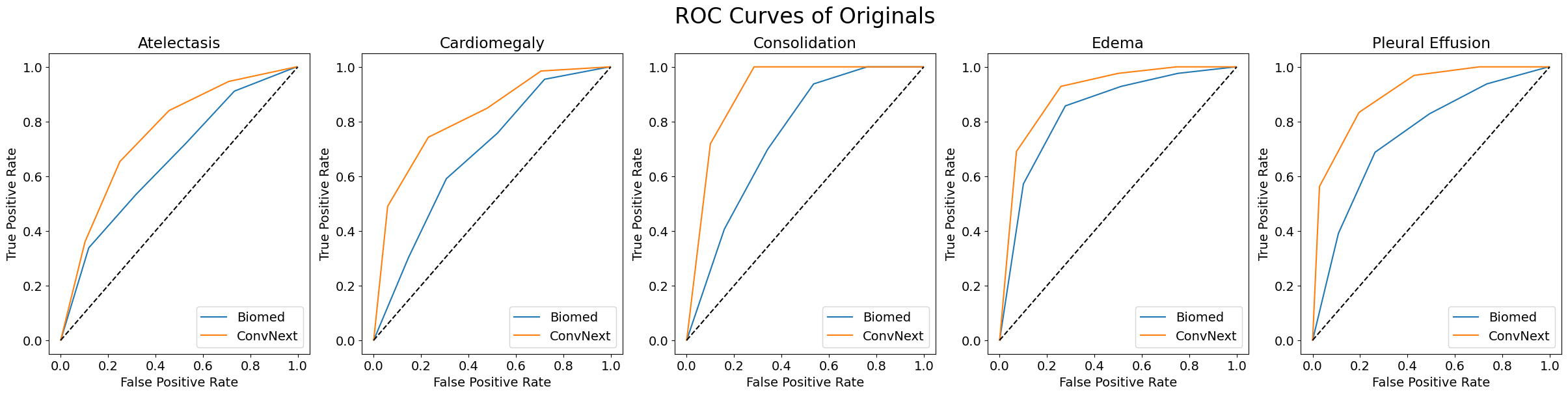}
    \includegraphics[width=\linewidth]{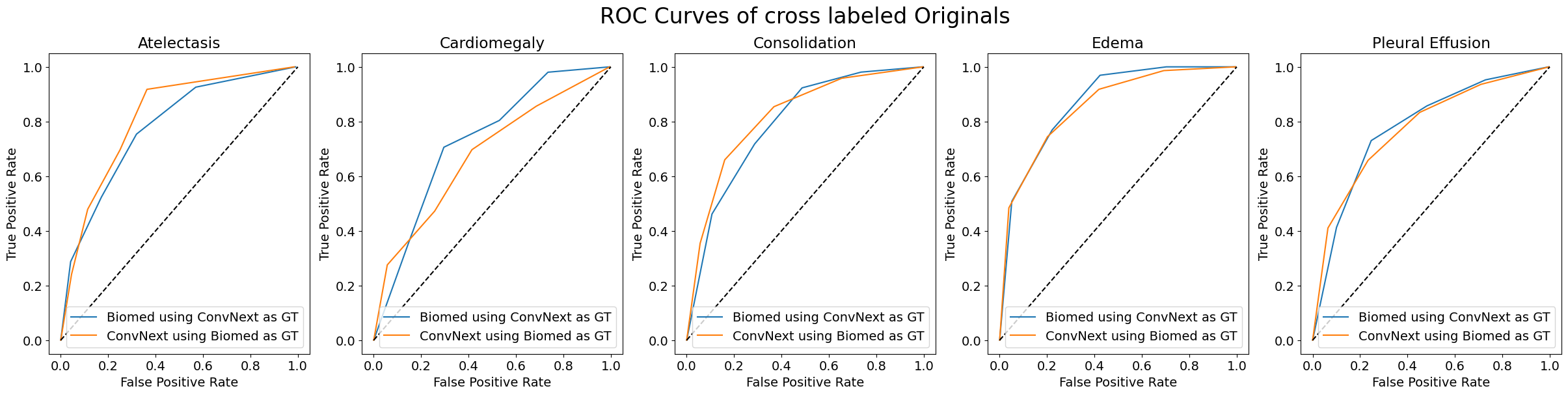}
    \includegraphics[width=\linewidth]{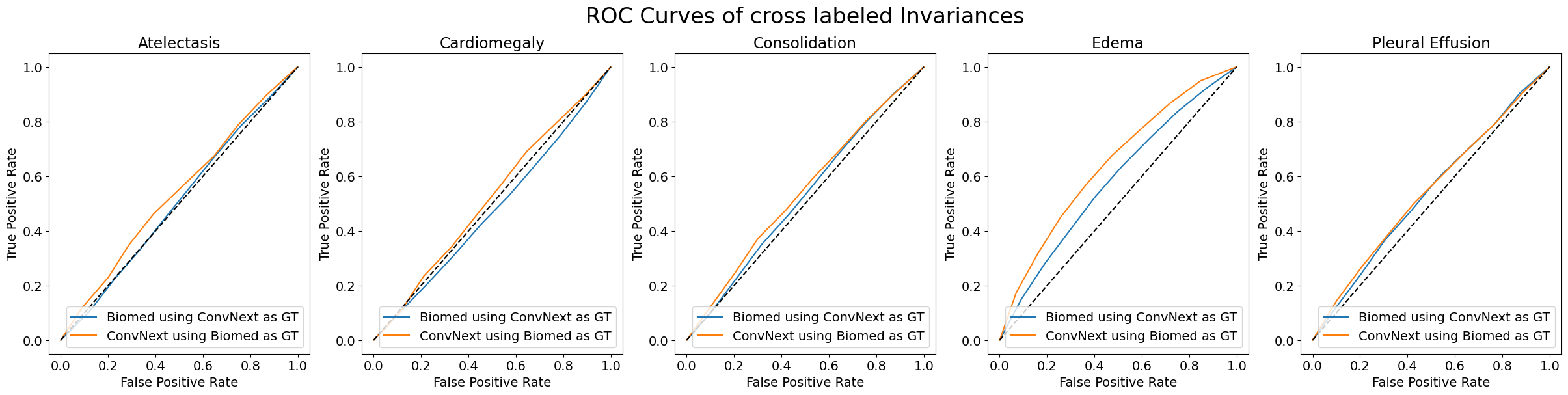}
    \caption{ROC curves for each of the five classes, that the CheXpert classifiers were trained on. We show results for comparing the BiomedClip classifier to the ConvNeXt classifier. Top: ROC curves for originals, using the ground truth labels. Center: ROC curves for originals, using the opposite model to create target labels. Bottom: ROC curves for invariances sampled for each model, with target labels created by the opposite model.}
    \label{fig:ChexpertROCCurves}
\end{figure}

In \cref{fig:ChexpertRandomSamples} we show random invariant samples of both subject models, compared to their nearest neighbors on the dataset. While the fiber loss for NDTM is the difference in logits, for a more interpretable distance measure we show fiber loss in terms of summed absolute difference in class probability $\sum\limits_{c=1}^{C} | p_c(x) - p_c(\tilde{x})|$.

We can see that NDTM achieves excellent fiber losses, with the assigned probabilities differing by less than $1\%$ in sum across all classes. In \cref{fig:ChexpertROCCurves} we highlight the difference in agreement between dataset originals and fiber samples for both models by showing ROC curves on various combinations of images and labels. When relabeling original images with either the ConvNeXt or BiomedClip model, we find their agreement rate on the original data (middle panel). We then label each fiber with the other subject model, then compute the ROC curves using the subject model with respect to which the fiber samples were originally drawn. This shows that their agreement mainly holds on original data, but not their fibers.

\begin{figure}[htb]
    \centering
    \includegraphics[width=\linewidth]{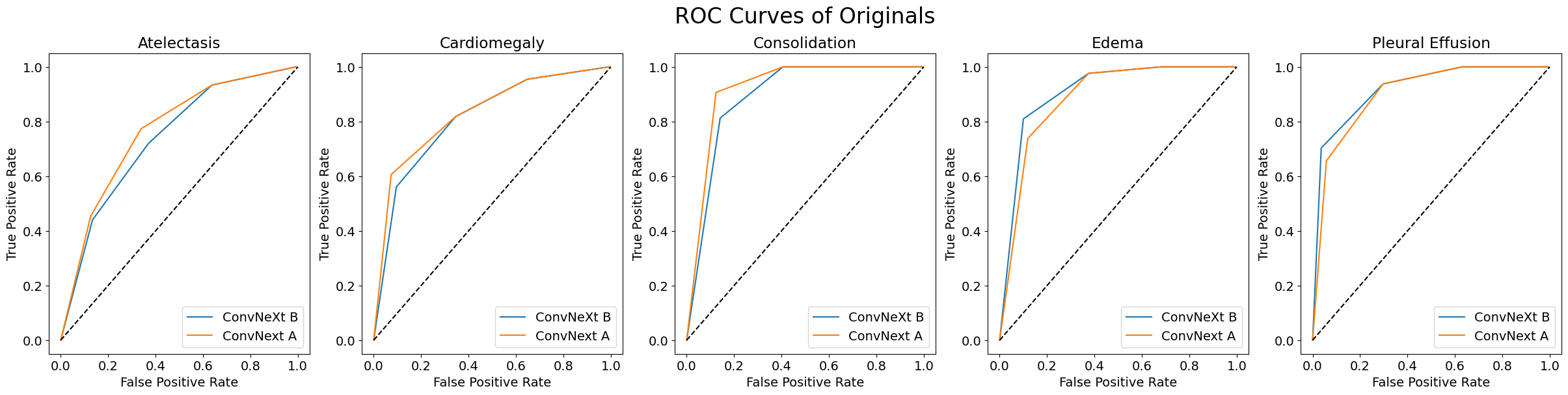}
    \includegraphics[width=\linewidth]{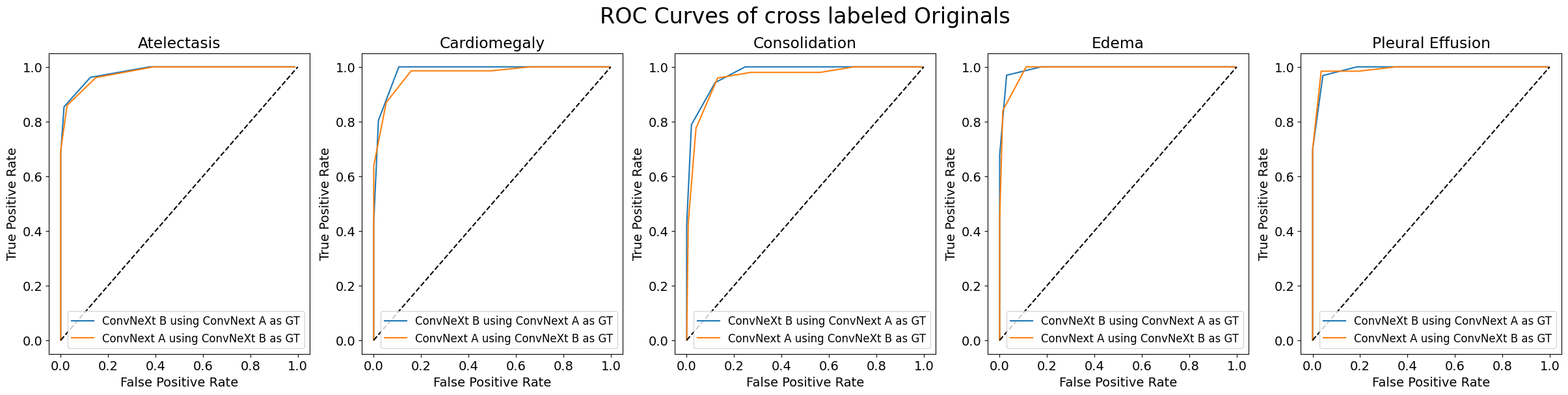}
    \includegraphics[width=\linewidth]{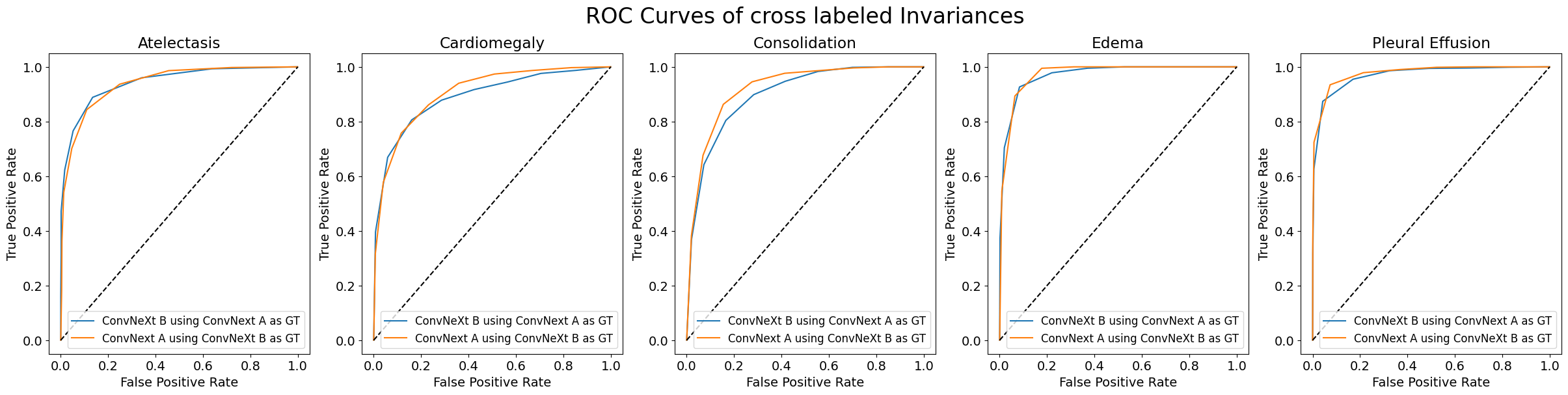}
    \includegraphics[width=0.7\linewidth]{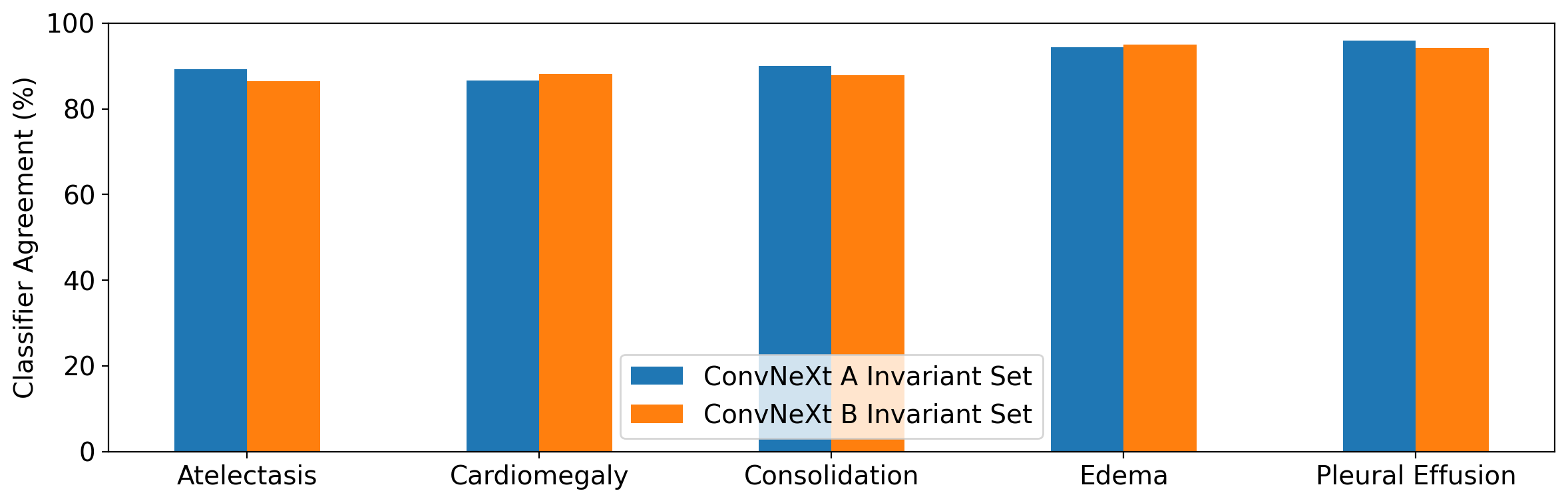}
    \caption{ROC curves for each of the five classes, that the CheXpert classifiers were trained on. We show results for two identically trained ConvNeXt classifiers. Top: ROC curves for originals, using the ground truth labels. Center: ROC curves for originals, using the opposite model to create target labels. Bottom: ROC curves for invariances sampled for each model, with target labels created by the opposite model. Below the ROC curves we show the classifier agreement on invariant sets sampled for originals, where both models predict the correct label.}
    \label{fig:ChexpertROCCurvesSame}
\end{figure}

Now we repeat the experiment for two ConvNeXt classifiers, which were trained identically. We show the corresponding ROC curves and classifier agreement measured analogous to \cref{sec:Chexpert} in \cref{fig:ChexpertROCCurvesSame}. This shows that identically trained models with the same architecture tend to also share the same fibers.

In order to verify that the invariances we find for the Qwen-2B vision encoder in rare anatomical cases are valid we can compare the fiber losses of \cref{fig:QwenExamples} with fiber losses on typical anatomy we obtained from a small sample of CheXpert representations. In the latter case the fiber loss is around $5.5 \pm 3.0$. The fiber losses of situs inversus images are comparable, meaning the fiber samples achieve the same level of invariance without reflecting the rare anatomy.

\subsection{Tabular Experiment: HTRU2}

To illustrate that invariance auditing can also be used for non-image modalities we audit the invariances of a tabular classifier on the HTRU2 dataset \citep{lyon2016fifty} for pulsar classification. The subject model classifier is a small MLP (2 layers, 128 hidden dimension) trained to detect pulsars from measured radio emissions. To perform fiber sampling we train an unconditional ResNet diffusion model (8 layers, 256 hidden dimension) and use the difference in classifier logits as fiber loss for NDTM.

\begin{figure}
    \centering
    \includegraphics[width=\linewidth]{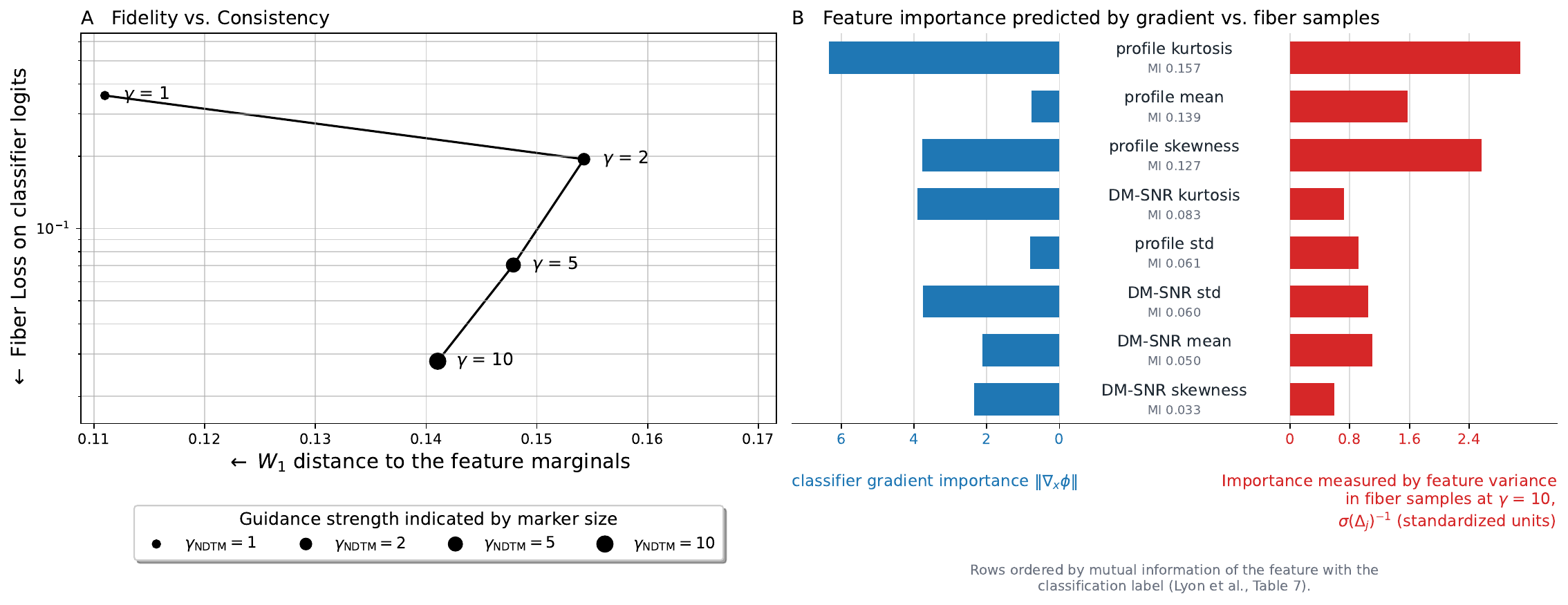}
    \caption{Invariance auditing of a tabular classifier on the HTRU2 dataset. Left: fidelity vs. consistency for fiber sampling with NDTM. The fiber losses correspond to differences in predicted probabilities of 6.1\% at $\gamma=1$ to 0.3\% at $\gamma=10$. Right: Feature importance as measured by gradients (blue) and feature variance in fiber samples (red). The features are ordered from the ones that have the most to least mutual information (MI) with the classification label.}
    \label{fig:htru2}
\end{figure}

First, we measure fidelity and consistency in \cref{fig:htru2} left, to verify that fiber sampling was successful. Then we use the fiber samples to determine the importance of each parameter in the following way: We normalize each input feature $j$ from the tabular dataset to obtain values $x_j$, then measure the standard deviation of the difference between the fiber sample value to the original value $\sigma(\Delta_j)$. The parameter importance is then the inverse of this:
\begin{align}
    \sigma(\Delta_j)^{-1} = \left(\sqrt{\mathbb{E}[|x_{j, \text{orig}}-x_{j, \text{fiber}}|^2] - \mathbb{E}[|x_{j, \text{orig}}-x_{j, \text{fiber}}|]^2}\right)^{-1}
\end{align}
The intuition behind this is that features that can vary a lot have low relevance for the classifier prediction, while parameters with small variation are important. In \cref{fig:htru2} right we compare the importance values obtained this way for each parameter to standard classifier gradient importance $\norm{\nabla_{x_j} \phi(x)}$. As a baseline we order features $j$ from most to least important based on the mutual information they have with the classification label (see Table 7 in \cite{lyon2016fifty}).

For all methods, profile kurtosis is the strongest predictor of pulsars, which is in line with it having the highest mutual information with the classification label. On the other features there is considerable disagreement. Gradient importance, which only considers variation in the local neighborhood, suggests that the classifier does not give higher importance to features with higher mutual information. Without profile kurtosis Spearman's rank correlation is $\rho = -0.04$ between the importance ranking of the gradient and mutual information. Invariance auditing, which considers the global variation of features, arrives at a different conclusion: The classifier's feature importance measured by fiber sample variance is strongly correlated with the importance ranking through mutual information, $\rho = 0.61$. It should be noted that, since the features in HTRU2 are themselves correlated, marginal feature importance measures have limited significance. Yet, this experiment highlights the ability of invariance auditing to perform global exploration of the fibers. Relying only on the local importance measures we may wrongly conclude that the classifier ignores high-information features, while invariance auditing reveals the correlation with mutual information on a global scale.

\section{Details on NDTM}
\label{app:CoarseNDTM}

In this section we provide additional details on NDTM and our modification of it.
NDTM guides the generation process of diffusion/flow-matching models, by adding a learnable correction $u_t$ between each denoising step
\begin{align}
    & \hat{x}_t = x_t + \gamma_t u_t \\
    & x_{t-1} = \text{DenoisingStep}(\hat{x}_t).
\end{align}
Here $\gamma_t$ is the guidance strength and $u_t$ is a parameter of same dimensionality as $x_t$, which is optimized to minimize the following loss function:
\begin{align}
    & x'_t \sim q(x'_t|x_0 = x) \\
    & \mathcal{L}_\text{terminal}(x_t, x'_t) = \left|\left| \phi(\mathbb{E}[x_0|x_t]) - \phi(\mathbb{E}[x_0|x'_t]) \right| \right|_2^2. \\
    & \mathcal{L}_\text{control} = \left|\left|u_t \right|\right|_2^2 \\
    & \mathcal{L}_\text{score} = \left| \left| \nabla_{x_t + \gamma_t u_t} \log p(x_t + \gamma_t u_t) -  \nabla_{x_t} \log p(x_t) \right| \right|_2^2\\
    & \mathcal{L} = \mathcal{L}_\text{terminal} + \kappa_t \mathcal{L}_\text{control} + \tau_t \mathcal{L}_\text{score}
    \label{eq:NDTM_loss}
\end{align}
$\mathcal{L}_\text{terminal}$ minimizes the terminal cost, in our case the fiber loss, $\mathcal{L}_\text{control}$ regularizes the size of the correction $u_t$ and $\mathcal{L}_\text{score}$ regularizes the change in score after the correction step, which is approximated by the diffusion network score estimate. We find the hyperparameter choices $\kappa_t = 10^{-4}$ and $\tau_t=0$ to be consistently stable choices for most settings. For each correction step, $u_t$ is initialized to zero and optimized by $N$ (usually 4 - 8) steps of gradient descent to minimize $\mathcal{L}$. The learning rate linearly decreases from $2\times10^{-3}$ at the start of the denoising process to $0$ at the end. Similar to classifier free guidance, we find it helpful to adjust the guidance strength $\gamma_t$ based on $t$. Specifically, we find that, depending on the diffusion model, high guidance strength is most helpful at a mid-range interval (starting around $t \in \left[ 800, 500 \right]$), similar to prior works \citep{kynkaanniemi2024applying}. Increasing $\gamma_t$ towards the end of the generation process can also help to reduce the fiber loss further, which we attribute to a better match in relevant fine textures.

\subsection{Proposed Modification}

\begin{figure}[htb]
    \centering
    \includegraphics[width=0.8\linewidth]{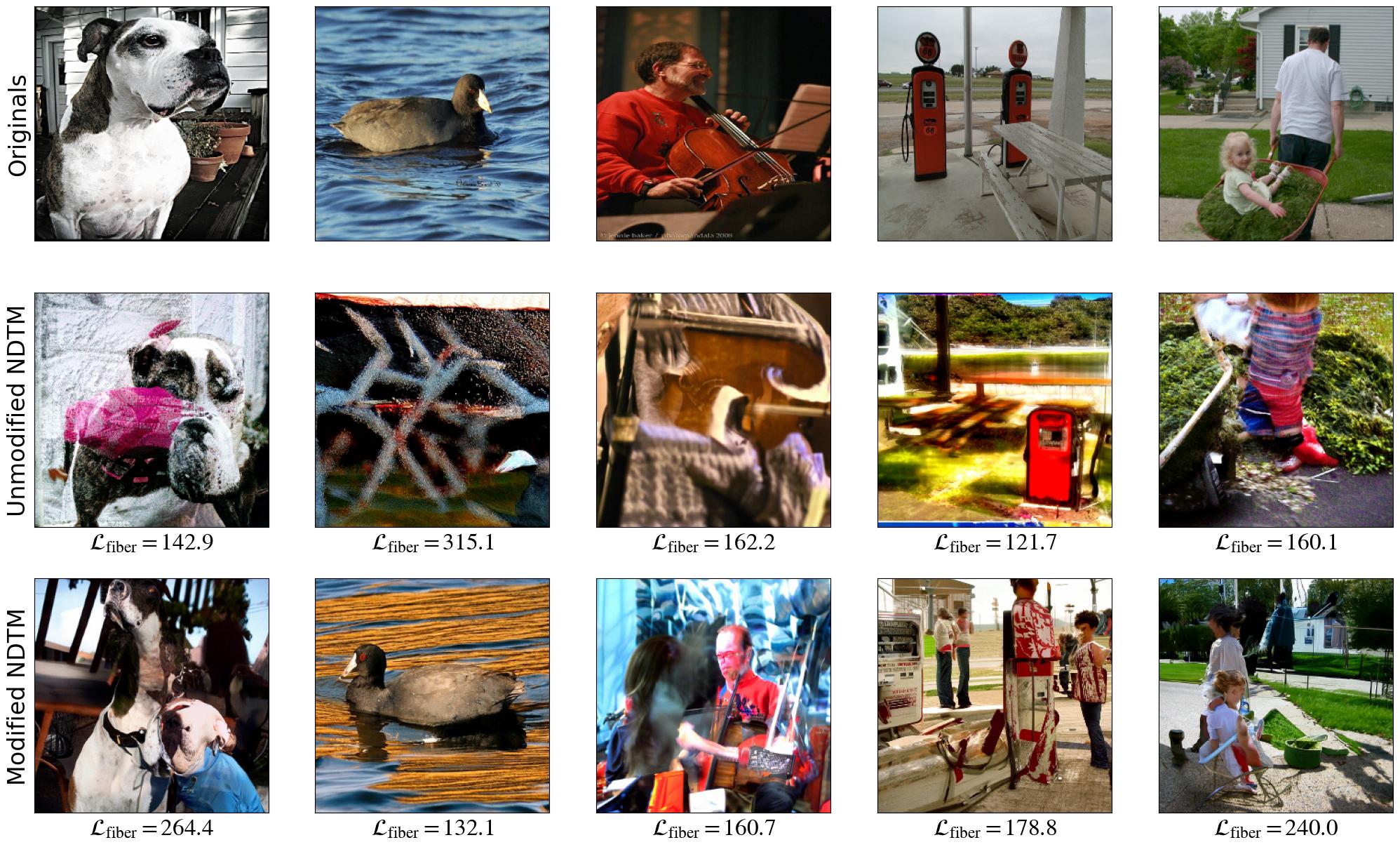}
    \caption{Random DINOv2 invariant samples on ImageNet without (middle panel), then with (lower panel) our modification to NDTM.}
    \label{fig:Static_vs_Moving}
\end{figure}

In \cref{sec:UncondFiberModel} we discussed our proposed modification to the terminal cost, that replaced the features of the input image $\phi(x)$ with that of a noised and re-approximated version $\phi(\mathbb{E}[x_0|x'_t])$. In \cref{fig:Static_vs_Moving} we show invariant samples of ImageNet taken with this new formulation compared to a static target. We can see that NDTM with re-approximated target features focuses a lot more on semantic content, while keeping colors and textures of the image similar to the underlying distribution $p(x)$. In contrast, static target feature guidance tries to create the same features that the detailed texture and colors of $x$ yield with the coarse approximation $\mathbb{E}[x_0|x_t]$, which results in out-of-distribution colors and textures in the final sample.  In \cref{tab:unmodified-vs-modified} we compare the resulting fiber loss and FID of the two sample sets. Our modification results in slightly higher fiber losses but increased stability (as indicated by the lower standard deviations of fiber losses) and slightly better consistency as measured by FID, which confirms its benefits for stability.

\begin{table}
    \centering
    \caption{Numerical results of performing invariance auditing for DINOv2 on ImageNet with unmodified and modified NDTM. Standard deviations are measured across the entire dataset.}
        \begin{tabular}{l|c|c}
    \toprule
        Metric & Unmodified NDTM & Modified NDTM \\ \midrule
        FID & 28.87 & 28.34 \\
        Fiber loss & $242 \pm 328$ & $272 \pm 142$ \\ \bottomrule
    \end{tabular}
    \label{tab:unmodified-vs-modified}
\end{table}

Empirically, we find that NDTM exhibits higher fiber losses in OOD settings (see \cref{tab:fiber-losses}) and on images-subject model combinations where representations depend on fine details that emerge late in the sampling process. In both cases we observe sampling instabilities for unmodified NDTM, which we suspect are due to a mismatch in granularity of the features present in the clean input image and the current noisy sample approximation. We find that our modification improves stability in these cases, but fiber loss is still elevated.

\begin{figure}[htb]
    \centering
    \includegraphics[width=\linewidth]{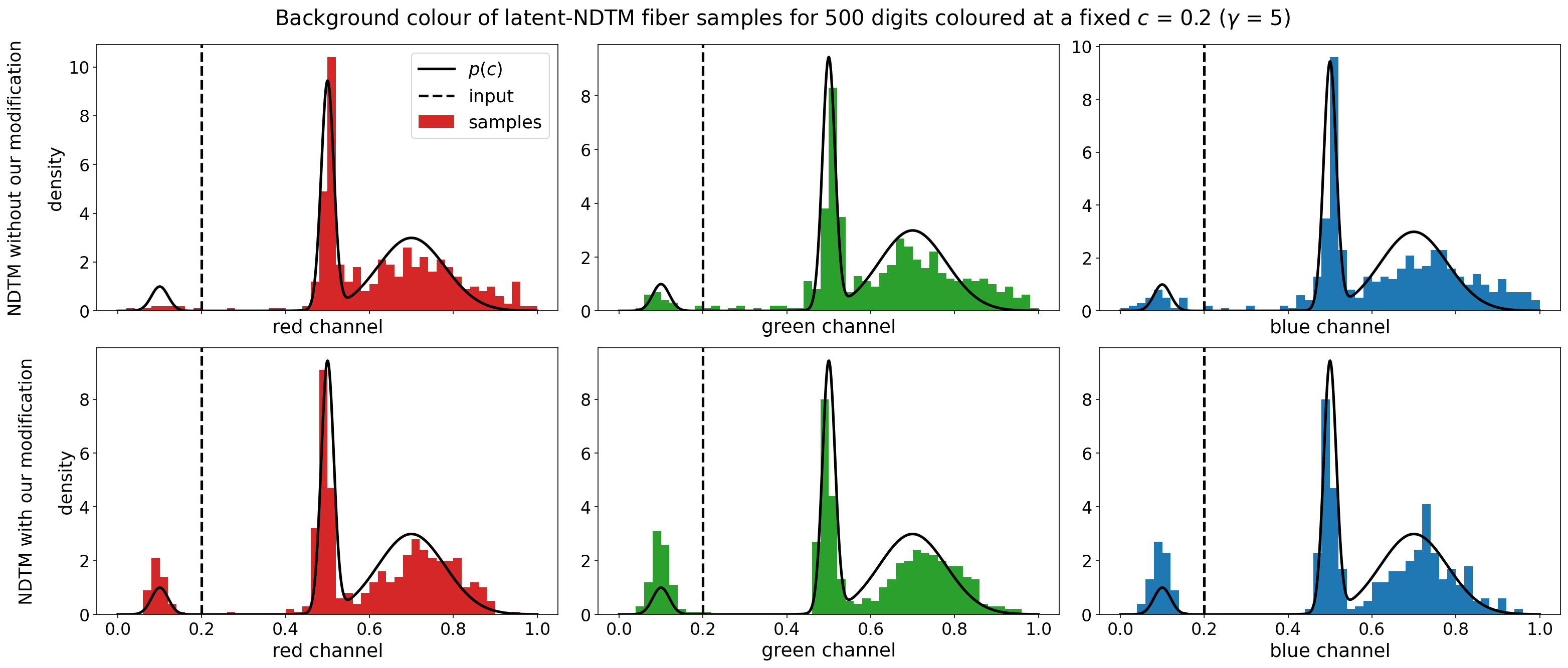}
    \caption{Color distribution resulting from coloring 5k MNIST digits at a fixed color (dashed line) and creating fiber samples with NDTM. The upper panel shows sampling without our modification, the lower panel with our modification. The solid black line shows the ground truth color distribution.}
    \label{fig:Static_vs_Moving_color_MNIST}
\end{figure}

As we use changing inputs to the subject model for modified NDTM, a possible concern is that additional information from $x$ may leak into the optimization process. To investigate this we include an additional experiment on color MNIST in \cref{fig:Static_vs_Moving_color_MNIST}, where we fix the background color (and by implication the digit color) to a constant value for 5k different images. Then we draw invariant samples using NDTM with and without our modification. If information could leak in the latter case, we should see a bias towards the fixed color, however, the sampled distribution $p_\theta(x|h)$ remains unchanged. Additionally, we can compare the diversity of our fiber samples for DINOv2 on ImageNet to those obtained by \citet{bordes2021high}. On a visual comparison, both methods manage to roughly reconstruct the same amount of information from the embeddings, verifying that our modification does not lead to a collapse onto the input image $x$.

\subsection{Choice of Fiber Loss Metric}
\label{app:MetricChoice}

In \cref{sec:Fibers} we introduced the fiber loss as the difference between target and fiber sample representations measured by any metric
\begin{align}
    \mathcal{L}_\text{fiber} = \text{dist}(\phi(x), \phi(\tilde{x})).
\end{align}

The fiber is fully determined by the subject model, and regardless of which metric we choose, if the fiber loss is zero, the sample will be exactly on the fiber. In practice, due to numerical limitations in the optimization process we only achieve fiber losses close to zero and the choice of fiber loss metric determines which shape of numerical deviations are preferred/tolerated. For this paper we chose to use $\ell_2$ distance as fiber loss metric, as it is a quite general choice in the absence of further information about the geometry of the subject model representations. As shown in \cref{tab:metric-comparison}, using $\ell_2$ fiber loss as terminal loss for NDTM also yields samples that have low distance as measured in other metrics -- as good as using these other metrics for NDTM themselves. We conclude that $\ell_2$ distance is a robust, geometry-agnostic choice. A more informed choice could be made on a case-by-case basis by taking into account the geometry of the subject model representations. Specifically, differences in representations could be measured by relevance to the task-outcome, e.g. using differences in classification probabilities for classifier subject models as was showcased for evaluation in \cref{app:Chexpert}. Since our aim in this paper is to be subject model agnostic wherever possible, we leave an exploration of different fiber loss metrics for future work.

\begin{table}
    \centering
    \caption{Comparison of different fiber loss metrics for invariance auditing of DINOv2 on ImageNet with NDTM. We use different metrics as terminal loss for NDTM (top to bottom) and report the fiber loss as measured by each metric (left to right). At the bottom we show the nearest neighbor distances for each metric in the dataset. Standard deviations are measured across the entire dataset.}
    \begin{tabular}{l|c|c|c}
    \toprule
    Terminal Loss Metric ↓ / Evaluation Metric → & $\ell_2$ & $\ell_1$ & Cross-Entropy \\
    \midrule
    $\ell_2$ & 272 ± 176 & 349 ± 103 & 5.34 ± 0.32 \\
    $\ell_1$& 277 ± 215 & 335 ± 110 & 5.36 ± 0.37 \\
    Cross-Entropy & 660 ± 461 & 533 ± 171 & 5.37 ± 0.29 \\ 
    Nearest Neighbor & 1224 ± 589 & 748 ± 186 & 5.17 ± 0.29 \\ \bottomrule
\end{tabular}

    \label{tab:metric-comparison}
\end{table}

\subsection{Ablation of NDTM hyperparameters}
\label{app:Ablation}

We perform additional ablations on NDTM hyperparameters on the color MNIST benchmark to determine the robustness of our results against hyperparameter choices. In particular, we show ablation results over the control and score regularizer strengths $\kappa$ and $\tau$, the sampling scheme used parameterized by $\eta$ (interpolating between DDIM at $\eta=0$ and DDPM at $\eta=1$), and finally the number of sampling steps. For the number of sampling steps in particular there are two variants: NDTM guidance is applied in each sampling step, so doubling the sampling steps effectively doubles the guidance. We therefore present two versions of this ablation, one where the guidance strength $\gamma$ is adjusted to compensate for this effect and one where it is kept constant. 

\begin{figure}
    \centering
    \includegraphics[width=0.49\linewidth]{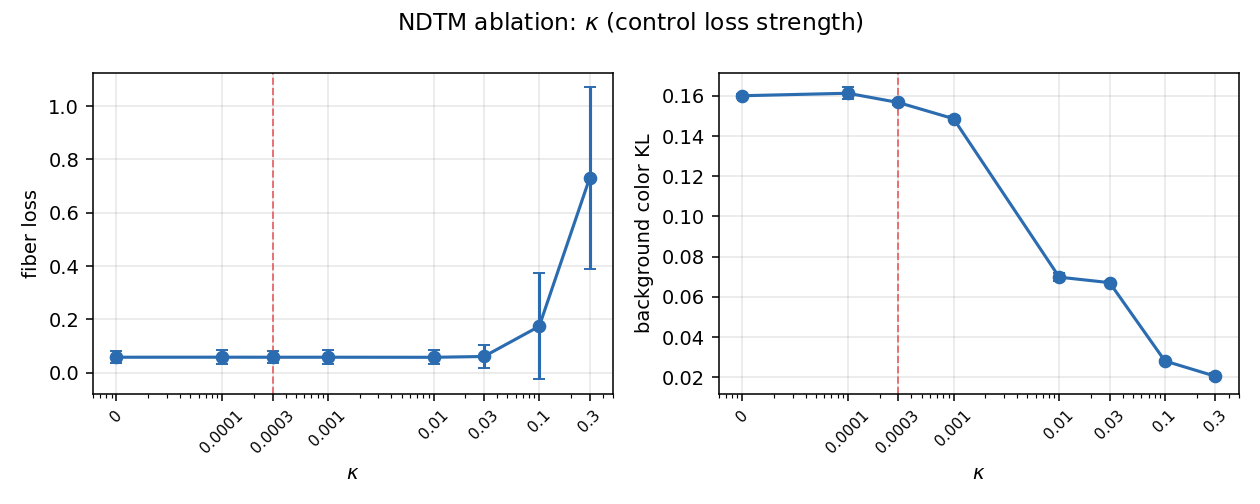}
    \includegraphics[width=0.49\linewidth]{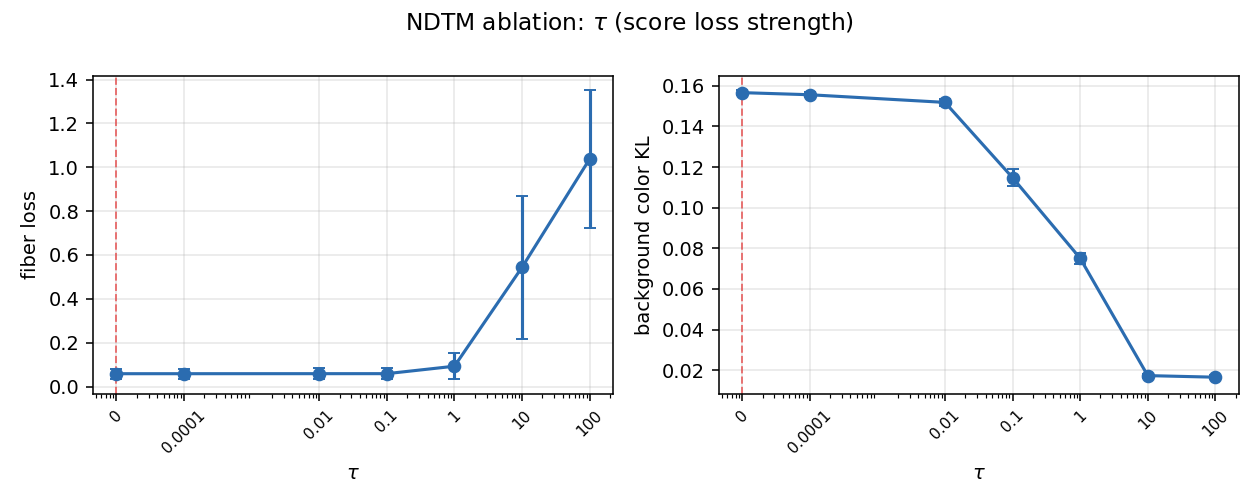}
    \includegraphics[width=0.49\linewidth]{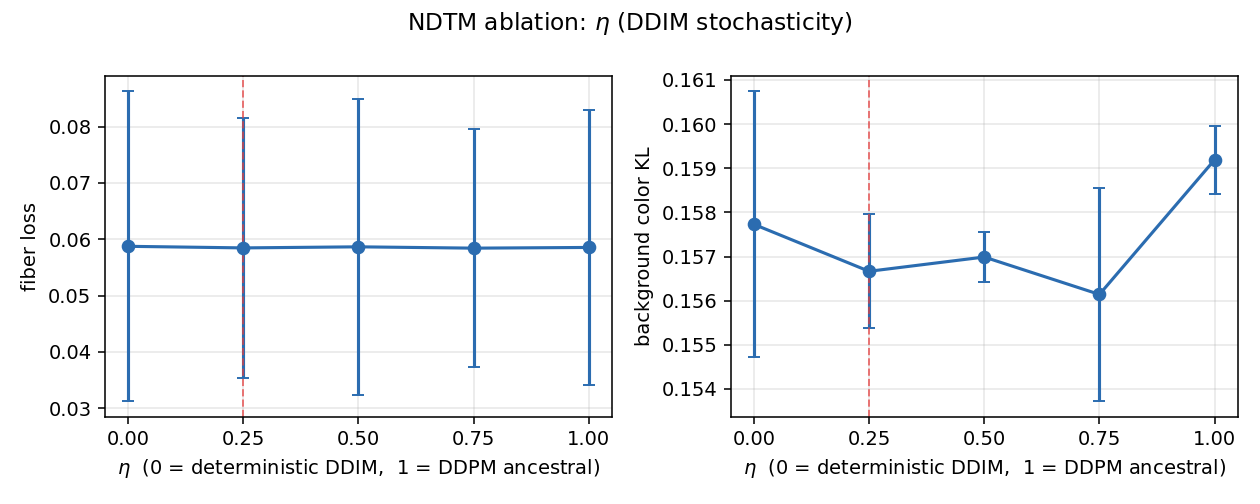}
    \includegraphics[width=0.49\linewidth]{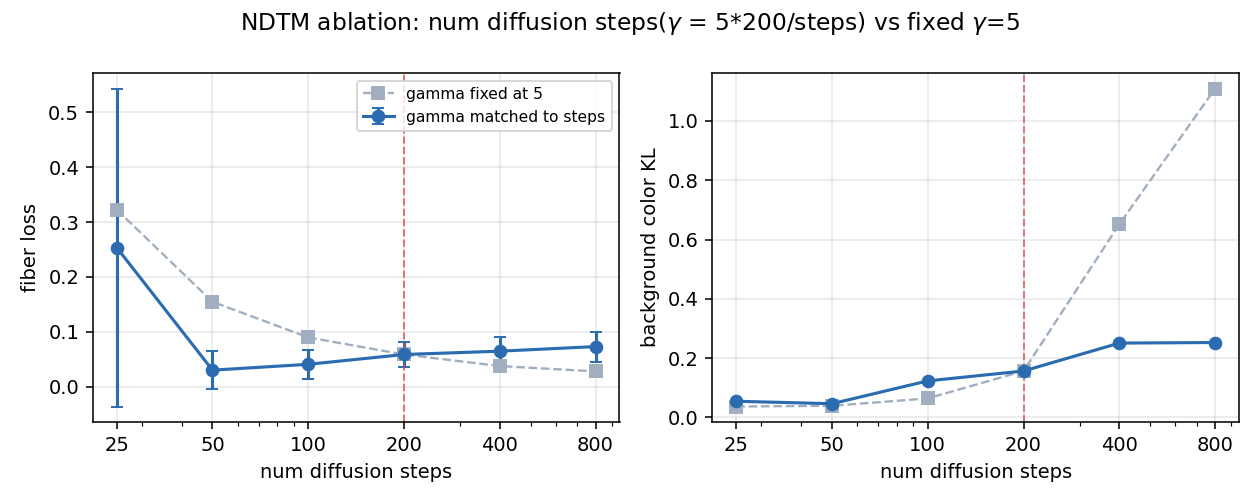}
    \caption{Ablations of different NDTM hyperparameters on color MNIST. Unless subject to ablation, all other parameters are kept at the values $\gamma$ = 5, $\kappa=3 \cdot 10^{-4}$, $\tau=0$, $\eta=0.25$, 200 diffusion steps, which were used for the experiments in \cref{sec:ColorMNIST} (marked by the red line).}
    \label{fig:Ablations}
\end{figure}

All results are shown in \cref{fig:Ablations}. We can see that NDTM for fiber sampling is robust against most hyperparameters: Sampler stochasticity $\eta$ has little to no effect, the regularizers both exhibit a long, stable plateau of no effect, then provide the expected trade-off between fidelity and consistency. We note \cref{fig:Ablations} suggests that for both regularizers there is a small interval where consistency can be significantly improved without significantly affecting fidelity until the trade-off begins. Finally, the number of sampling steps shows the expected fidelity-consistency trade-off through larger effective guidance when not compensating through $\gamma$. With the compensation, we find that a low number of sampling steps can lead to even better results in both than the 200 we used for \cref{fig:ColorMNISTPareto}.

\begin{table}
    \centering
    \caption{Licenses of used assets}
    \resizebox{\linewidth}{!}{
        \begin{tabular}{lll}
\toprule
\textbf{Asset} & \textbf{Type} & \textbf{License} \\
\midrule

MNIST (LeCun et al., 2010) & Dataset & CC BY-SA 3.0 \\
ImageNet & Dataset & Custom (non-commercial / restricted) \\
CheXpert (Irvin et al., 2019) & Dataset & Custom Stanford license (non-commercial) \\
Cue Conflict Dataset (Geirhos et al., 2018) & Dataset & CC BY 4.0 \\
Color MNIST (Rombach et al., 2020) & Dataset & CC BY-SA 3.0 (derived) \\

\midrule

DINOv2 ViT-B/14 (Oquab et al., 2023) & Model & Apache 2.0 \\
InceptionV3 (Szegedy et al., 2016) & Model & Apache 2.0 \\
ResNet-50 (ImageNet pretrained) & Model & BSD 3-Clause \\
ConvNeXt (Liu et al., 2022) & Model & MIT \\
BiomedCLIP (Zhang et al., 2023) & Model & MIT \\
Qwen-2B vision encoder & Model & Custom Apache 2.0 \\
ImageNet diffusion model (Dhariwal and Nichol, 2021) & Model & MIT \\

\midrule

Kaggle notebook (Gaddam, 2024) & Code & Apache 2.0 \\

\bottomrule
\end{tabular}

    }
    \label{tab:assets}
\end{table}

\begin{table}
    \centering
    \caption{Total estimated GPU compute time of each experiment, including training of fiber and subject models where no pretrained models are used. The main source of compute cost was sampling 10k fiber samples to compute FID for ImageNet models. All experiments were performed on a compute cluster with NVIDIA A100 40GB GPUs.}
    \begin{tabular}{ccc}
    \toprule
    \shortstack{Experiment} & \shortstack{Total GPU time}  \\    
    \midrule
    ColorMNIST & $\sim 100$h \\
    ImageNet & $\sim 800$h \\
    Cue Conflict & $\sim 400$h \\
    CheXpert & $\sim 140$h \\
    Causal MNIST & $< 5$h \\
    HTRU2 & $<5$h \\
    \bottomrule
\end{tabular}
    \label{tab:experiment-runtimes}
\end{table}

\clearpage
\section*{NeurIPS Paper Checklist}

\begin{enumerate}

\item {\bf Claims}
    \item[] Question: Do the main claims made in the abstract and introduction accurately reflect the paper's contributions and scope?
    \item[] Answer: \answerYes{} 
    \item[] Justification: For each of our main claims we provide a matching experiment, specifically the performance and scalability of guided invariance auditing is shown on colorMNIST and ImageNet.

\item {\bf Limitations}
    \item[] Question: Does the paper discuss the limitations of the work performed by the authors?
    \item[] Answer: \answerYes{} 
    \item[] Justification: We discuss limitations in \cref{sec:Conclusion}, including the training-guiding compute trade-off and the need for methods to remove blind spots once they have been identified.

\item {\bf Theory assumptions and proofs}
    \item[] Question: For each theoretical result, does the paper provide the full set of assumptions and a complete (and correct) proof?
    \item[] Answer: \answerNA{} 
    \item[] Justification: The paper does not include theoretical results.

    \item {\bf Experimental result reproducibility}
    \item[] Question: Does the paper fully disclose all the information needed to reproduce the main experimental results of the paper to the extent that it affects the main claims and/or conclusions of the paper (regardless of whether the code and data are provided or not)?
    \item[] Answer: \answerYes{} 
    \item[] Justification: We provided details on all experimental setups in \cref{app:ExperimentalDetails} and the pretrained models used are all publicly available. We will release our code upon acceptance.

\item {\bf Open access to data and code}
    \item[] Question: Does the paper provide open access to the data and code, with sufficient instructions to faithfully reproduce the main experimental results, as described in supplemental material?
    \item[] Answer: \answerNo{} 
    \item[] Justification: We will release our code upon acceptance.

\item {\bf Experimental setting/details}
    \item[] Question: Does the paper specify all the training and test details (e.g., data splits, hyperparameters, how they were chosen, type of optimizer) necessary to understand the results?
    \item[] Answer: \answerYes{} 
    \item[] Justification: Experimental details are given in \cref{app:ExperimentalDetails}. Each experiment section in the main contains the high-level experimental setting which is necessary to understand the results.

\item {\bf Experiment statistical significance}
    \item[] Question: Does the paper report error bars suitably and correctly defined or other appropriate information about the statistical significance of the experiments?
    \item[] Answer: \answerYes{} 
    \item[] Justification: We provide standard deviations for all numerical results in our tables. Qualitative results in \cref{sec:Chexpert} are explicitly labeled as such.

\item {\bf Experiments compute resources}
    \item[] Question: For each experiment, does the paper provide sufficient information on the computer resources (type of compute workers, memory, time of execution) needed to reproduce the experiments?
    \item[] Answer: \answerYes{} 
    \item[] Justification: We provide approximate runtimes of all experiments in \cref{tab:experiment-runtimes}.
    
\item {\bf Code of ethics}
    \item[] Question: Does the research conducted in the paper conform, in every respect, with the NeurIPS Code of Ethics \url{https://neurips.cc/public/EthicsGuidelines}?
    \item[] Answer: \answerYes{} 
    \item[] Justification: None of the ethical concerns listed in the code of ethics apply in our case.

\item {\bf Broader impacts}
    \item[] Question: Does the paper discuss both potential positive societal impacts and negative societal impacts of the work performed?
    \item[] Answer: \answerNA{} 
    \item[] Justification: With this work our goal is to advance the interpretability and trustworthiness of machine learning models. We do not anticipate any negative societal impacts from this.
    
\item {\bf Safeguards}
    \item[] Question: Does the paper describe safeguards that have been put in place for responsible release of data or models that have a high risk for misuse (e.g., pre-trained language models, image generators, or scraped datasets)?
    \item[] Answer: \answerNA{} 
    \item[] Justification: Our paper does not present any new models/data that has high risk of misuse. 

\item {\bf Licenses for existing assets}
    \item[] Question: Are the creators or original owners of assets (e.g., code, data, models), used in the paper, properly credited and are the license and terms of use explicitly mentioned and properly respected?
    \item[] Answer: \answerYes{} 
    \item[] Justification: All existing assets are properly credited via citation. Licenses are provided in \cref{tab:assets}.

\item {\bf New assets}
    \item[] Question: Are new assets introduced in the paper well documented and is the documentation provided alongside the assets?
    \item[] Answer: \answerYes{} 
    \item[] Justification: We introduce a new benchmark in \cref{sec:ColorMNIST}, which is documented in \cref{sec:app_cmnist}.

\item {\bf Crowdsourcing and research with human subjects}
    \item[] Question: For crowdsourcing experiments and research with human subjects, does the paper include the full text of instructions given to participants and screenshots, if applicable, as well as details about compensation (if any)? 
    \item[] Answer: \answerNA{} 
    \item[] Justification: No human subjects were involved in this paper.

\item {\bf Institutional review board (IRB) approvals or equivalent for research with human subjects}
    \item[] Question: Does the paper describe potential risks incurred by study participants, whether such risks were disclosed to the subjects, and whether Institutional Review Board (IRB) approvals (or an equivalent approval/review based on the requirements of your country or institution) were obtained?
    \item[] Answer: \answerNA{} 
    \item[] Justification: No human subjects were involved in this paper.

\item {\bf Declaration of LLM usage}
    \item[] Question: Does the paper describe the usage of LLMs if it is an important, original, or non-standard component of the core methods in this research? Note that if the LLM is used only for writing, editing, or formatting purposes and does \emph{not} impact the core methodology, scientific rigor, or originality of the research, declaration is not required.
    \item[] Answer: \answerNA{} 
    \item[] Justification: LLMs do not constitute any core, non-standard components of our method development.

\end{enumerate}

\end{document}